\documentclass[10pt,twocolumn,letterpaper]{article}

\usepackage[pagenumbers]{cvpr} 

\usepackage{cuted}
\usepackage{booktabs}
\usepackage{makecell}

\usepackage{xspace}
\newcommand{\methodname}{EmbroMaster\xspace}
\newcommand{\datasetname}{StitchAtlas\xspace}

\definecolor{cvprblue}{rgb}{0.21,0.49,0.74}
\usepackage[pagebackref,breaklinks,colorlinks,allcolors=cvprblue]{hyperref}
\usepackage{titletoc}
\usepackage{etoc}

\def\confName{CVPR}
\def\confYear{2027}

\title{\methodname: Controllable Multi-Stitch Embroidery Synthesis with Craft-Aware Planning}

\author{
Zhanlin Chen$^{1,*}$,
Zheng Gu$^{2,*}$,
Hao Liang$^{1}$,
Wenhua Shi$^{3,\dagger}$,
Xin Jin$^{4,5}$,
Jing Huo$^{1,\dagger}$,
Yang Gao$^{1}$\\
$^{1}$State Key Laboratory for Novel Software Technology, Nanjing University\\
$^{2}$College of Computer Science and Software Engineering, Shenzhen University\\
$^{3}$School of Art and Design, Shanghai Institute of Technology\\
$^{4}$Beijing Electronic Science and Technology Institute\\
$^{5}$State Key Laboratory of General Artificial Intelligence,\\
Beijing Institute for General Artificial Intelligence (BIGAI)\\
}

\begin{document}
\maketitle
\begingroup
\renewcommand{\thefootnote}{\fnsymbol{footnote}}
\footnotetext[1]{Equal contribution}
\footnotetext[2]{Corresponding author: huojing@nju.edu.cn}
\endgroup


\etocdepthtag.toc{main}

\begin{strip}
    \vspace{-35pt}
    \centering
    \includegraphics[width=\textwidth]{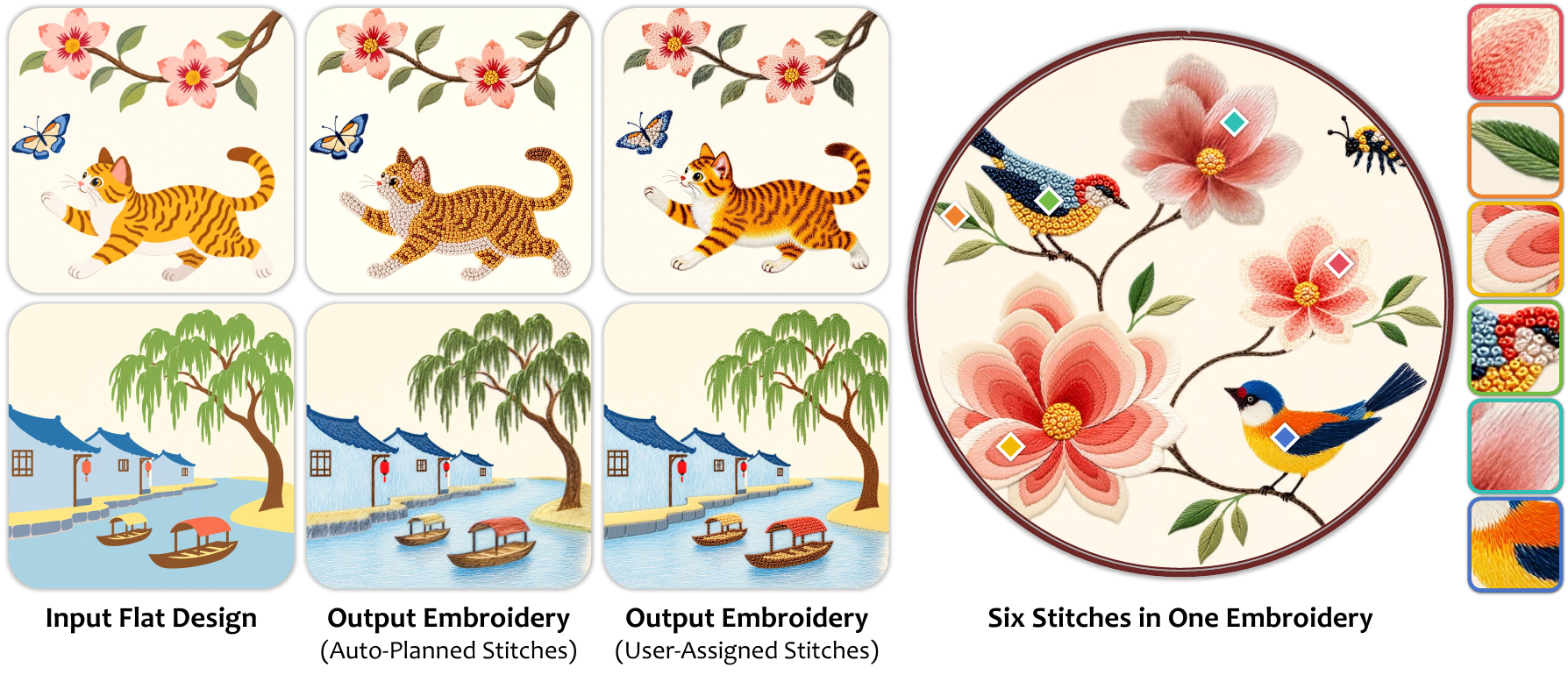}
    \captionsetup{hypcap=false}
    \captionof{figure}{
        \textbf{Controllable multi-stitch embroidery with \methodname.}
        Left: Embroideries generated from flat designs with automatically
        planned or user-assigned stitches. Right: A single embroidery
        combining six stitch types, with details of thread textures.
    }
    \label{fig:teaser}
\end{strip}
\begin{abstract}
Generating embroidery takes more than imitating thread texture: a model must decide which stitch suits each region and render several distinct stitches together in one image.
However, current generative models lack the embroidery knowledge to plan which stitch a region should use.
Even when specific rules are given, they still struggle to render multiple stitches in one image.
In this paper, we present \methodname to address these problems.
We first construct \datasetname, a multi-stitch embroidery dataset. We collect real embroidery samples and train models on them to expand the dataset with high-quality synthetic images, along with stitch craft knowledge collected from embroidery experts.
We then introduce a craft-aware planning agent. It takes a flat design and user constraints as input and produces a structured stitch plan, called an embroidery blueprint, that assigns a stitch to each region.
Finally, we develop a multi-stitch synthesis pipeline that follows the blueprint to generate embroidery images and supports interactive, region-level stitch editing.
Experiments on six stitch types demonstrate that our method achieves better generation quality than existing methods.
Embroidery experts also rate our results higher on stitch assignment appropriateness and visual realism, suggesting the potential of our method for various application scenarios.
Our project page is available at https://njuczl.github.io/EmbroMaster/.
\end{abstract}
    
\section{Introduction}
\label{sec:intro}

Su embroidery is a traditional Chinese craft distinguished by delicate stitch techniques~\cite{yang2018paint}. Its appearance depends on matching semantic regions with specialized stitches, each producing distinct thread structures and visual effects.
Traditional workflows of embroidery manufacturing require artisans to analyze input patterns, plan stitch arrangements for individual regions, and refine the results through trial-and-error stitching, which is time-consuming and relies heavily on expert experience.

\begin{figure}[t]
    \centering
    \includegraphics[trim=163 218 160 162, clip, width=\columnwidth]{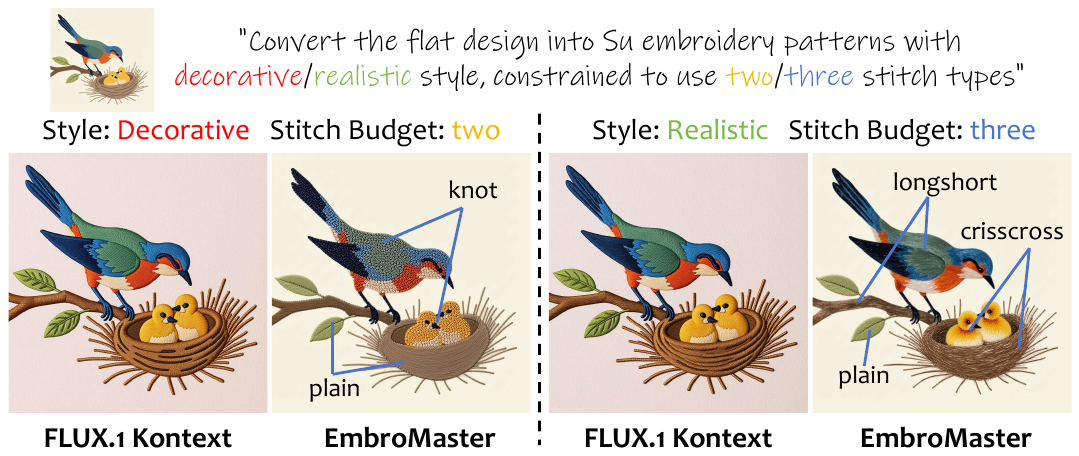}
    \caption{Given an input flat design, FLUX.1 Kontext~\cite{batifol2025fluxkontext} produces undifferentiated textures. \methodname generates multi-stitch Su embroidery following the constraints.}
    \label{fig:kontextvsembromaster}
\end{figure}

Recent advances in generative models have greatly improved image synthesis abilities~\cite{zhang2023controlnet,brooks2023instructpix2pix,esser2024rectifiedflow}.
However, three challenges remain for multi-stitch embroidery synthesis.
\textbf{First}, real Su embroidery images with specific stitch types are scarce. Beyond images, the craft knowledge that governs stitch assignment remains implicit in expert experience, which has not been explicitly investigated.
\textbf{Second}, even if the craft knowledge is given, applying it requires joint reasoning over semantic content and embroidery expression. 
Existing multimodal models can recognize objects but lack the craft-aware planning ability to determine which stitch should be applied to each region.
\textbf{Third}, given a stitch assignment, current image generation methods struggle to produce realistic multi-stitch embroidery with fine-grained details, and do not support consistent and controllable region-level editing.

To address these challenges, our idea is to separate the decision of \textit{which stitch should be used where} from the rendering of \textit{how each stitch should appear}. We decompose the embroidery synthesis into two stages: \textbf{craft-aware planning} and \textbf{multi-stitch synthesis}.
Bridging these two stages, we leverage a structured representation, which we call an embroidery blueprint, to explicitly record the semantic region, spatial mask, and assigned stitch for every region in the design.
The blueprint is first generated by a craft-aware planning agent that reasons over user input and stitch craft knowledge, then serves as an editable interface for controllable multi-stitch embroidery synthesis.

In the first planning stage, the agent identifies semantic embroidery units, such as petals or leaves, and assigns stitches according to stitch craft knowledge and user constraints, including style and stitch budget. It then invokes a segmentation model to obtain spatial masks, completing the blueprint. Users can revise stitch assignments, with the agent updating only the affected decisions. In the second synthesis stage, stitch-specific renderers trained with LoRA stylize the full input design, while the blueprint's masks select the regions to retain. We compose these regions in decreasing order of mask area so that larger background regions do not overwrite smaller semantic details.

To enable the above framework, we construct \datasetname, a multi-stitch embroidery dataset combining embroidery images with stitch craft knowledge. For stitch types where real samples are scarce, we train a LoRA model separately to produce diverse synthetic embroidery images. We convert real and synthetic images into flat designs by removing textures while preserving object boundaries and colors, forming pixel-aligned design--stitch pairs. Experts manually filter artifacts and misaligned samples to obtain high-quality pairs for training stitch-specific renderers. We also collect craft rules from Su embroidery experts on how to select stitches based on semantic regions, visual structures, and desired artistic effects. These rules guide the planning agent towards better stitch assignments.

We evaluate \methodname on six representative Su embroidery stitch types.
Experiments demonstrate that our method achieves the best performance in rule compliance and generation quality.
Notably, we conduct a user study with ten Su embroidery experts, who rate our results highest across five dimensions, showcasing the practical value of integrating craft knowledge and explicit planning into computational embroidery design.
\Cref{fig:teaser} shows examples of multi-stitch synthesis and region-level editing with EmbroMaster.
\Cref{fig:kontextvsembromaster} further illustrates \methodname's ability to generate distinct stitch appearances while respecting region-level stitch assignments.

Our contributions are summarized as follows:

\begin{itemize}
\item We formulate multi-stitch embroidery synthesis as a plan-and-generate problem and introduce a craft-aware planning agent, providing an explicit and explainable framework for multi-stitch design and assignment.
\item We develop a multi-stitch synthesis pipeline with stitch-specific renderers and area-ordered composition. Our pipeline generates realistic multi-stitch embroidery and supports consistent region-level editing.
\item We construct a multi-stitch embroidery dataset, \datasetname, which expands sparse real embroidery samples into design--stitch pairs, together with stitch craft knowledge collected from embroidery experts.
\end{itemize}

\begin{figure*}[t]
    \centering
    \includegraphics[width=\textwidth]{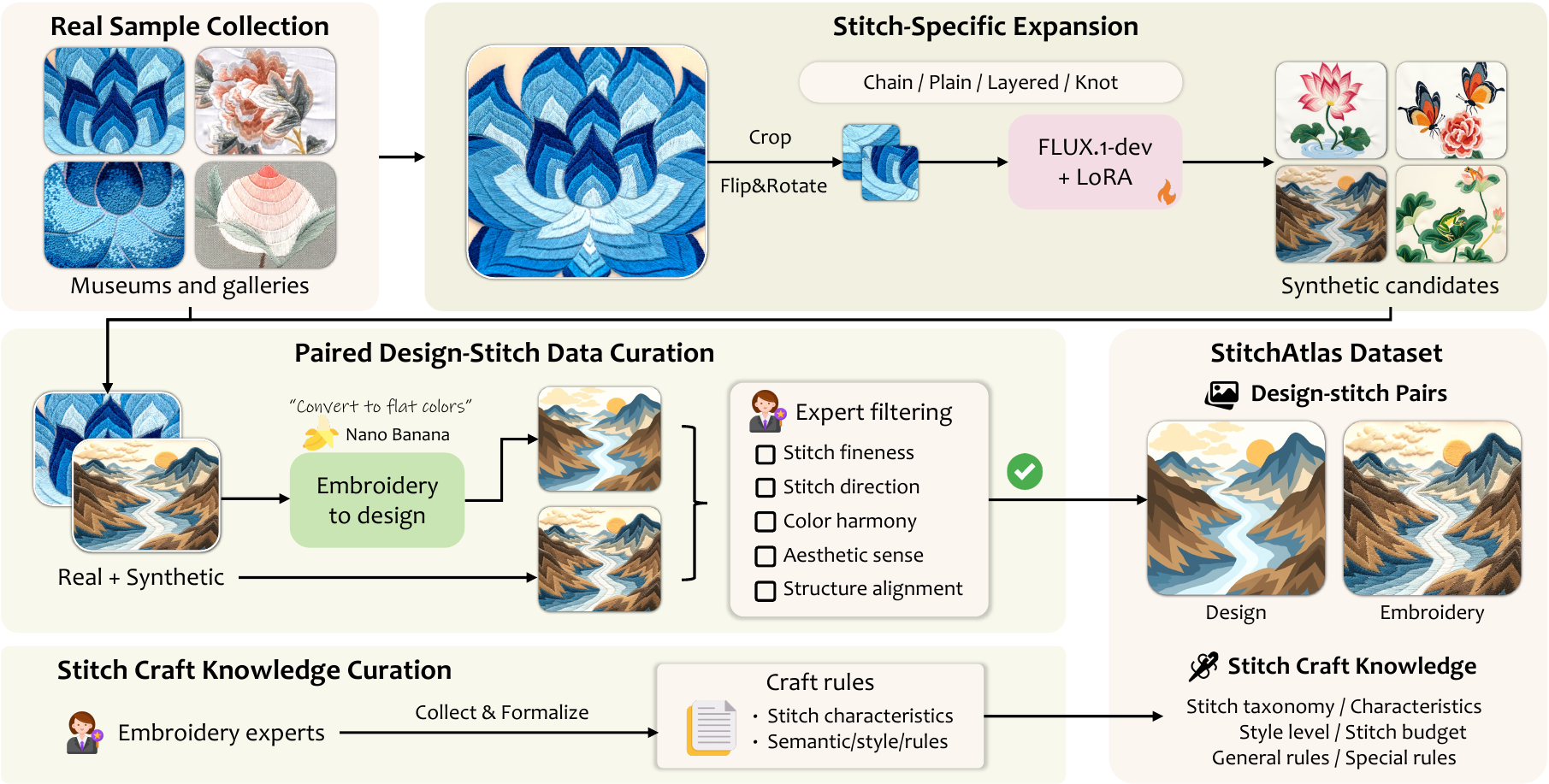}
    \caption{\textbf{\datasetname dataset construction.} 
    Real sample collection and stitch-specific expansion provide real and synthetic embroideries. Paired design-stitch data curation converts them to flat-color designs and uses expert filtering to select high-quality aligned design-stitch pairs for training stitch-specific renderers. In parallel, stitch craft knowledge curation formalizes expert rules for craft-aware planning.
    }
    \label{fig:dataset_pipeline}
\end{figure*}

\section{Related Work}
\subsection{Image Stylization and Texture Synthesis}

Classical texture synthesis methods construct appearance from example patches~\cite{efros2001quilting,hertzmann2001analogies}, while neural style transfer methods typically match global feature statistics or transform feature distributions~\cite{gatys2016style,huang2017adain,li2017wct}. Recent diffusion-based methods improve controllability through attention sharing~\cite{hertz2024stylealigned}, training-free style injection~\cite{chung2024styleinjection}, reference-based customization~\cite{jones2024paircustomization}, fine-grained style--content modulation~\cite{frenkel2024blora}, and structural conditioning~\cite{zhang2023controlnet}. However, these methods target appearance transfer or spatial control, rather than assigning distinct textures to semantic regions according to domain-specific knowledge. In contrast, \methodname explicitly plans stitch assignments based on semantic content and stitch craft knowledge, then composes stitch-specific renderings into a coherent embroidery image.

\subsection{Computational Embroidery Modeling}
Early computational embroidery methods explore procedural stitch modeling, image-based rendering, and sample-based region filling~\cite{yang2018paint,chen2012embroidery,yang2012irregular,ma2022multilayer,guan2021automatic}. Subsequent work investigates paired embroidery-style generation~\cite{ye2021endtoend}, unsupervised generation with embroidery-aware attention~\cite{yang2022channelattention}, and directionality-aware pattern design~\cite{liu2023directionality}. Recent approaches include MSEmbGAN~\cite{hu2025msembgan} for region-aware multi-stitch synthesis and EmoLoRA~\cite{ma2025emolora} for one-shot embroidery customization through contrastive LoRA modulation.
MSEmbGAN assigns three predefined stitch types to color regions using region-width heuristics and synthesizes their textures with a shared stitch-conditioned network. Its assignment mechanism relies on region geometry rather than explicit semantic and craft-aware reasoning, and incorporating additional stitch types would require adapting its assignment scheme and texture-generation model. EmoLoRA instead focuses on one-shot customization of an embroidery style, rather than explicit region-level stitch planning and multi-stitch composition. In contrast, \methodname integrates expert-provided stitch craft knowledge with semantic reasoning to produce an editable embroidery blueprint, and uses stitch-specific renderers trained on real and synthetic design--stitch pairs. This modular design supports region-level stitch editing and the incorporation of new stitches through additional craft rules and a new renderer, without retraining existing renderers. Additional comparisons with EmoLoRA are provided in the supplementary material.

\subsection{Large Vision-Language Models in Visual Tasks}
Large language and vision-language models have demonstrated strong capabilities in semantic reasoning and compositional spatial planning~\cite{feng2023layoutgpt,lai2024lisa}, while segmentation foundation models enable open-vocabulary region extraction~\cite{carion2026sam3}. Building on these capabilities, \methodname employs a vision-language model as a planning agent rather than a monolithic image generator. The agent assigns stitches to semantic regions according to stitch craft knowledge and user preferences, while a segmentation model provides spatial masks and specialized renderers synthesize stitch appearances. This modular design combines domain-specific reasoning with controllable visual generation.

\begin{figure*}[t]
    \centering
    \includegraphics[width=\textwidth]{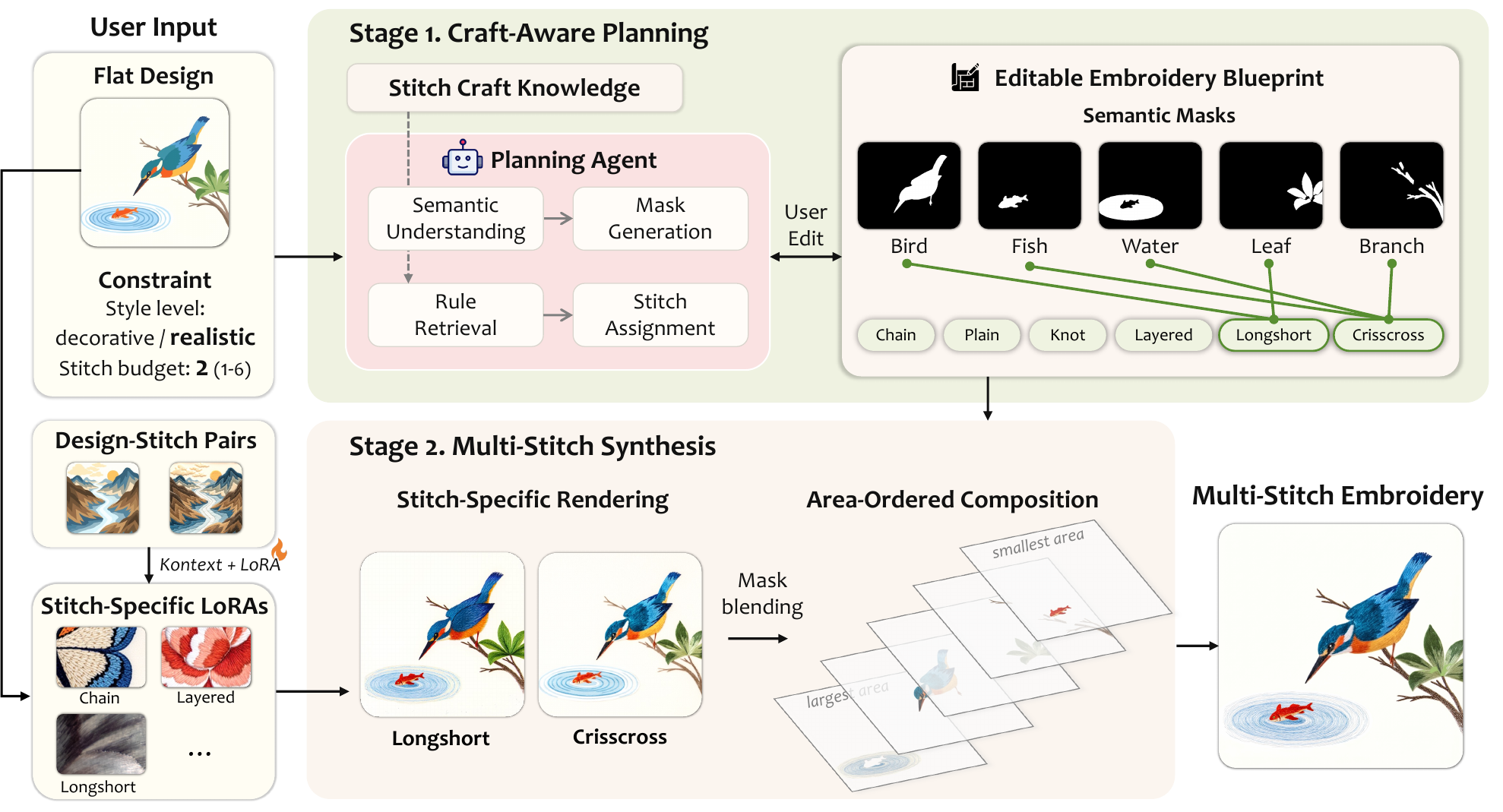}
    \caption{\textbf{Overview of \methodname.} 
    In Stage 1 (craft-aware planning), the planning agent uses stitch craft knowledge to translate a flat design and user constraints into an editable embroidery blueprint specifying semantic masks and stitch assignments. Stage 2 (multi-stitch synthesis) uses these assignments to select only the required stitch-specific LoRAs trained on design-stitch pairs for global stitch-specific rendering. Area-ordered composition uses the blueprint masks to blend regions from large to small, producing multi-stitch embroidery.
    }
    \label{fig:framework}
\end{figure*}

\section{Method}
\label{sec:method}

\subsection{Framework Overview}
\label{sec:problem}
We first construct the \textbf{\datasetname} dataset (\Cref{sec:data}, \Cref{fig:dataset_pipeline}), which provides design--stitch pairs for training stitch-specific renderers and stitch craft knowledge for planning. As shown in \Cref{fig:framework}, \methodname follows two stages connected by an editable embroidery blueprint. In \textbf{craft-aware planning} (\Cref{sec:planning}), the planning agent uses this knowledge to generate semantic masks and stitch assignments from a flat design and user constraints, including style level and stitch budget. In \textbf{multi-stitch synthesis} (\Cref{sec:generation}), the selected stitch-specific renderers generate full-image single-stitch renderings, which are combined according to the blueprint through area-ordered composition.

\subsection{\datasetname Dataset Construction}
\label{sec:data}

Our stitch vocabulary $\mathcal{S}$ comprises six representative stitch types: chain, plain, layered, knot, long-short, and crisscross.
As shown in \Cref{fig:dataset_pipeline}, we construct design--stitch pairs through real sample collection, stitch-specific expansion, and paired data curation, and collect and formalize stitch craft knowledge with embroidery experts.

\paragraph{Paired Design-Stitch Data Curation.}
We collect high-resolution Su embroidery samples from museums and galleries, with fewer than five clean samples available for many stitches and no paired flat-color designs.
For these data-scarce stitches, we augment the real samples through cropping, flipping, and rotation, then train stitch-specific LoRAs~\cite{hu2022lora} on FLUX.1-dev~\cite{blackforestlabs2024flux1} to generate additional single-stitch embroidery candidates.
We use Nano Banana Pro~\cite{google2025nanobananapro} (abbreviated as Nano Banana in \Cref{fig:dataset_pipeline}) to convert real and synthetic embroideries into flat-color designs, removing textures while preserving structures and major color regions.
Each converted design is paired with its source embroidery: the design serves as the training input and the source embroidery as the target.
Experts filter these candidate design--stitch pairs based on stitch fineness, stitch direction, color harmony, aesthetic sense, and structure alignment.
For each stitch $s\in\mathcal{S}$, we retain 20 high-quality pairs as $\mathcal{D}_s$ to train the renderers in \Cref{sec:generation}. 

\paragraph{Stitch Craft Knowledge Curation.}
We collect and formalize knowledge from embroidery experts, covering stitch taxonomy and characteristics, style level and stitch budget, and general and special rules.
\textit{Stitch taxonomy} distinguishes point-based (knot), line-based (chain), and surface-based (plain, layered, long-short, and crisscross) techniques.
\textit{Stitch characteristics} describe their thread structures, textures, and typical uses.
\textit{Style level} specifies decorative or realistic expression, while \textit{stitch budget} specifies the exact number of distinct stitch types to use.
\textit{General rules} associate semantic content, style, and region scale with candidate stitches.
\textit{Special rules} prioritize expert preferences, such as knot for flower stamens and long-short or crisscross for animal fur.
The full vocabulary, style descriptions, and rules are provided in the supplementary material.

\subsection{Craft-Aware Planning}
\label{sec:planning}

\paragraph{Craft-Aware Planning Agent.}
Given a flat design $D$ and user constraints, the planning agent uses the stitch craft knowledge from \datasetname to construct the blueprint through four related functions.
\textbf{Semantic understanding} identifies embroidery units from their semantic content and region scale.
\textbf{Mask generation} invokes SAM 3~\cite{carion2026sam3} to obtain the corresponding binary masks.
\textbf{Rule retrieval} selects applicable rules from the provided stitch craft knowledge according to the design semantics and user constraints.
\textbf{Stitch assignment} combines these rules with the semantic analysis to assign one stitch from $\mathcal{S}$ to each unit, while matching the requested style level and stitch budget.
The masks and stitch assignments together form the blueprint
\begin{equation}
    \mathcal{B}=\{(o_i,M_i,t_i)\}_{i=1}^{n},
    \qquad t_i\in\mathcal{S},
    \label{eq:blueprint}
\end{equation}
where $n$ is the number of embroidery units, $o_i$ denotes the semantic region of unit $i$, $M_i$ is its binary spatial mask, and $t_i$ is its assigned stitch.

\paragraph{Editable Embroidery Blueprint.}
The blueprint's \textit{stitch assignments} determine which stitch-specific renderers are used, while its \textit{masks} specify where their outputs contribute to the final image.
Users can inspect the object--stitch mapping and revise local assignments or planning preferences.
The planning agent then updates the affected decisions, and the multi-stitch synthesis pipeline follows the revised blueprint without retraining the renderers.

\subsection{Multi-Stitch Synthesis}
\label{sec:generation}

We realize the embroidery blueprint through stitch-specific rendering followed by area-ordered composition.

\paragraph{Stitch-Specific Rendering.}
For each stitch $s\in\mathcal{S}$, we train a separate stitch-specific LoRA for FLUX.1-Kontext-dev~\cite{batifol2025fluxkontext} on $\mathcal{D}_s$, yielding an image-to-image renderer $\mathcal{G}_s$.
At inference, we invoke only the renderers selected by the blueprint, each of which processes the full design $D$ once to produce a global single-stitch rendering:
\begin{equation}
    I_s=\mathcal{G}_s(D).
    \label{eq:stitch_renderer}
\end{equation}
Each renderer specializes in one stitch, reducing interference between stitch characteristics.
The blueprint masks then select the assigned regions from these global renderings, preserving the design's overall structure and color organization while allowing local stitch variation.
A new stitch can be introduced by adding its paired data, renderer, and craft rules without retraining the existing renderers.

\paragraph{Area-Ordered Composition.}
In overlapping or nested regions, compositing a large region after a smaller one may overwrite details such as eyes or stamens.
We therefore use binary mask area as a proxy for semantic scale and compose regions from large to small, giving smaller semantic details priority in these regions.
Let $\sigma$ be a permutation of the $n$ units such that
\begin{equation}
    |M_{\sigma(1)}|\geq\cdots\geq|M_{\sigma(n)}|,
    \label{eq:area_order}
\end{equation}
where $|M_i|$ is the number of foreground pixels in mask $M_i$.
For blending, we apply Gaussian feathering to each binary mask $M_i$ to obtain a soft mask $\alpha_i$ with values in $[0,1]$.
Starting from the original design as the initial canvas, $I^{(0)}=D$, we recursively compose
\begin{equation}
    \begin{aligned}
        I^{(k)}={}I^{(k-1)}\odot(1-\alpha_{\sigma(k)})
        +I_{t_{\sigma(k)}}\odot\alpha_{\sigma(k)},
    \end{aligned}
    \label{eq:composition}
\end{equation}
for $k=1,\ldots,n$, where $\odot$ denotes elementwise multiplication with the blending weights applied to every color channel, and $I_{t_{\sigma(k)}}$ is the global rendering for the assigned stitch.
The final embroidery is $I_{\mathrm{final}}=I^{(n)}$.
Pixels with zero blending weight in every mask retain the original design.
This area ordering governs image composition and does not specify a physical stitching sequence.

\section{Experiments}
\label{sec:experiments}

\subsection{Experimental Setup}
\label{sec:exp_setup}

\paragraph{Implementation Details.}
We evaluate on 200 multi-object flat designs with decorative or realistic style and an exact stitch budget of two or three. 
For craft-aware planning, the planning agent uses GPT-6 Sol~\cite{openai2026gpt6sol} for reasoning and SAM 3 for mask generation.
We implement the stitch-specific renderers in \Cref{sec:generation} using separate FLUX.1-Kontext-dev LoRAs trained on the corresponding paired datasets $\mathcal{D}_s$.
For these renderer LoRAs, we set both the rank and LoRA alpha to 32, use a learning rate of $1\times10^{-4}$, and train each renderer for 2,000 steps.
All LoRA models are trained on an NVIDIA A6000 GPU.
For area-ordered composition, we encode each binary mask as 0 or 255, apply a $9\times9$ Gaussian blur, and divide the blurred values by 255 to obtain the soft blending weights $\alpha_i$ used in \Cref{eq:composition}.
More details are provided in the supplementary material.

\begin{figure*}[t]
    \centering
    \includegraphics[trim=0 5 0 0, clip,width=\textwidth]{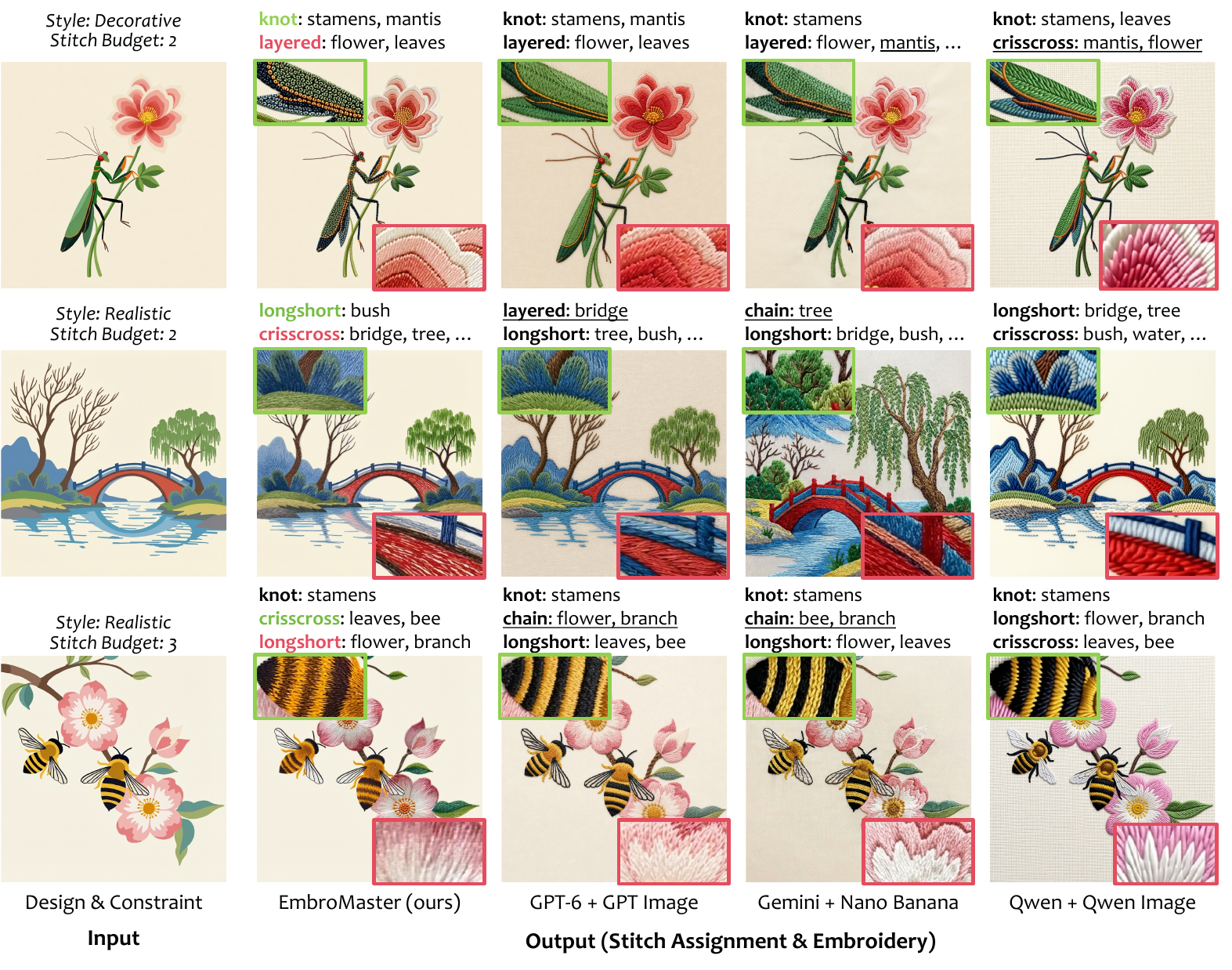}
    \caption{\textbf{Qualitative comparisons with baseline methods.}
    Given the same inputs, EmbroMaster, GPT, Gemini, and Qwen generate stitch assignments and embroideries. Boxes show enlarged details. \underline{Underlined} entries indicate inappropriate stitch assignments.}
    \label{fig:qualitative_multi}
\end{figure*}

\begin{table*}[t]
    \centering
    \caption{\textbf{Quantitative comparisons with baseline methods.}
    Best and second-best results are highlighted in bold and underlined, respectively.
    LPIPS, RSP, Hist. Loss, and L-KID are scaled by $100$; L-HFRD by $1000$.}
    \label{tab:overall_comparison}

    \small
    \setlength{\tabcolsep}{3pt}
    \renewcommand{\arraystretch}{1.15}

    \begin{tabular*}{\textwidth}{
        @{\extracolsep{\fill}}lcccccccc@{}
    }
        \toprule
        Method
        & LPIPS $\downarrow$
        & RSP $\uparrow$
        & Hist. Loss $\downarrow$
        & L-$\Delta E_{00}$ $\downarrow$
        & L-HFRD $\downarrow$
        & L-KID $\downarrow$
        & Stitch Acc. $\uparrow$
        & Rule Comp.$\uparrow$ \\
        \midrule

        Qwen
        & 45.676
        & 36.817
        & 64.290
        & 18.734
        & 7.947
        & 14.738
        & 0.180
        & \underline{0.642} \\

        GPT
        & \underline{21.094}
        & 40.919
        & \underline{49.932}
        & \underline{18.062}
        & 8.195
        & 10.980
        & 0.300
        & 0.617 \\

        Gemini
        & 31.643
        & \underline{45.660}
        & 63.348
        & 19.682
        & \underline{7.387}
        & \textbf{9.037}
        & \underline{0.339}
        & 0.630 \\

        EmbroMaster
        & \textbf{11.319}
        & \textbf{57.411}
        & \textbf{32.653}
        & \textbf{15.063}
        & \textbf{4.163}
        & \underline{9.214}
        & \textbf{0.712}
        & \textbf{0.975} \\

        \bottomrule
    \end{tabular*}
\end{table*}

\paragraph{Evaluation Metrics.}
We evaluate the results along four dimensions: 
\textbf{perceptual and structural fidelity}, measured by LPIPS ($\downarrow$)~\cite{zhang2018lpips} and Region-wise Structure Preservation (RSP, $\uparrow$), computed using masked low-pass SSIM~\cite{wang2004ssim}; 
\textbf{color fidelity}, measured by Histogram Loss (Hist. Loss, $\downarrow$)~\cite{aviaharon2023histogram} and Local Color $\Delta E_{00}$ (L-$\Delta E_{00}$, $\downarrow$)~\cite{luo2001ciede2000}; 
\textbf{texture and distributional similarity}, measured by Local High-Frequency Ratio Difference (L-HFRD, $\downarrow$)~\cite{ma2025emolora} and Local KID (L-KID, $\downarrow$)~\cite{binkowski2018kid}; 
and \textbf{stitch correctness}, measured by Stitch Classification Accuracy (Stitch Acc., $\uparrow$) and Stitch Assignment Rule Compliance (Rule Comp., $\uparrow$). 
More details are provided in the supplementary material.

\begin{figure*}[t]
    \centering
    \includegraphics[width=\textwidth]{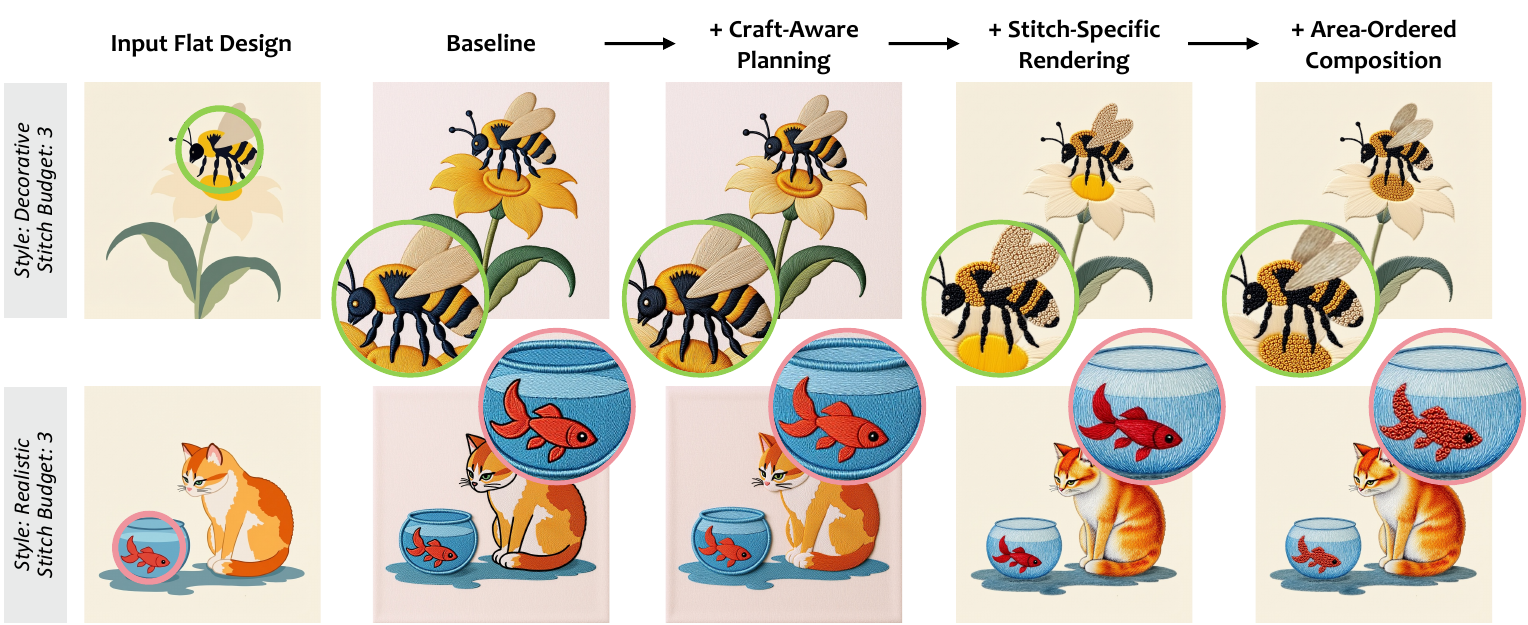}
    \caption{\textbf{Qualitative ablation study.}
    Starting from the baseline, we progressively add craft-aware planning, stitch-specific rendering, and area-ordered composition to synthesize embroideries from the input designs. The circles show enlarged details of each result.}
    \label{fig:ablation}
\end{figure*}

\begin{table*}[t]
\centering
\caption{\textbf{Quantitative ablation study.}
CAP denotes craft-aware planning. SSR denotes stitch-specific rendering. AOC denotes area-ordered composition.
Best and second-best results are highlighted in bold and underlined, respectively.
}
\label{tab:progressive_ablation}

\small
\setlength{\tabcolsep}{3pt}
\renewcommand{\arraystretch}{1.15}

\begin{tabular*}{\textwidth}{
    @{\extracolsep{\fill}}lcccccccc@{}
}
    \toprule
    Configuration
    & LPIPS $\downarrow$
    & RSP $\uparrow$
    & Hist. Loss $\downarrow$
    & L-$\Delta E_{00}$ $\downarrow$
    & L-HFRD $\downarrow$
    & L-KID $\downarrow$
    & Stitch Acc. $\uparrow$ 
    & Rule Comp. $\uparrow$ \\
    \midrule

    Baseline
    & 23.494
    & 39.521
    & 69.809
    & 25.019
    & 11.866
    & 15.516
    & 0.280 
    & \underline{0.617} \\

    Baseline + CAP
    & 23.524
    & 39.667
    & 69.562
    & 25.032
    & 13.354
    & 15.321
    & 0.192 
    & \textbf{0.975} \\

    Baseline + CAP + SSR
    & \textbf{11.250}
    & \textbf{58.225}
    & \textbf{32.647}
    & \textbf{14.914}
    & \underline{4.412}
    & \textbf{9.211}
    & \underline{0.698} 
    & \textbf{0.975} \\

    Baseline + CAP + SSR + AOC
    & \underline{11.319}
    & \underline{57.411}
    & \underline{32.653}
    & \underline{15.063}
    & \textbf{4.163}
    & \underline{9.214}
    & \textbf{0.712} 
    & \textbf{0.975} \\

    \bottomrule
\end{tabular*}
\end{table*}

\subsection{Comparison with Baseline Methods}
\label{sec:baseline_comparison}

We compare \methodname with three general-purpose multimodal generation pipelines: Gemini (Gemini 3.8 Flash + Nano Banana Pro)~\cite{google2025nanobananapro,google2026gemini38flash}, GPT (GPT-6 Sol + GPT-Image-2)~\cite{openai2026gpt6sol,openai2026gptimage2}, and Qwen (Qwen3.8-Flash-Next + Qwen-Image-Edit-2511)~\cite{qwen2026qwen38flashnext,qwen2025qwenimageedit2511,wu2025qwenimage}. All methods perform stitch planning followed by multi-stitch synthesis. All methods receive the same inputs: the design, user constraints on style and stitch budget, stitch references and a fixed object list shared across methods.
Stitch references show the characteristic appearance of each stitch type to the baseline methods, while using the same fixed object list controls for differences in object identification across methods.

\paragraph{Qualitative Comparisons.}
\label{sec:qualitative}

In \Cref{fig:qualitative_multi}, the competing methods exhibit both inappropriate stitch assignments (underlined) and inconsistencies between the assigned and rendered stitches.
In the first row, GPT assigns knot stitch to the mantis but fails to reproduce the characteristic texture of knot stitches.
Gemini assigns layered stitch to the same object, yet renders a granular texture.
These examples highlight the difficulty of translating a stitch plan into the corresponding visual appearance.
Gemini also changes the input content in the second row, while Qwen produces uniform, low-quality stitch textures across the entire image.
\methodname better preserves the depicted content and reproduces the assigned stitch appearances in these examples.

\paragraph{Quantitative Comparisons.}
\label{sec:quantitative}

\methodname ranks first on seven of the eight metrics in \Cref{tab:overall_comparison}, including Stitch Acc.\ of $0.712$ and Rule Comp.\ of $0.975$.
It ranks second on L-KID ($9.214$), behind Gemini ($9.037$).
These results indicate strong design preservation, faithful execution of stitch assignments, and adherence to stitch craft knowledge.

\subsection{Ablation Study}
\label{sec:ablation}
Our ablation baseline combines GPT-6 Sol for planning with FLUX.1-Kontext-dev for image synthesis, without stitch craft knowledge or LoRA fine-tuning.
We progressively add craft-aware planning, stitch-specific rendering, and area-ordered composition, as shown in \Cref{fig:ablation} and \Cref{tab:progressive_ablation}.
Without area-ordered composition, the stitch-specific renderings are composited in a random region order.

\paragraph{Craft-Aware Planning.}
Adding craft-aware planning increases Rule Comp.\ from $0.617$ to $0.975$. However, improved planning compliance alone does not ensure high-quality stitch textures from the unadapted image model, as reflected in limited visual improvement in \Cref{fig:ablation} and the lack of consistent gains across other metrics in \Cref{tab:progressive_ablation}.

\paragraph{Stitch-Specific Rendering.}
Introducing the stitch-specific renderers raises Stitch Acc.\ to $0.698$ and substantially improves structure, color, and texture metrics, providing the largest gains in generation quality.
The examples in \Cref{fig:ablation} show more distinctive stitch textures while retaining the design's structure and colors.

\paragraph{Area-Ordered Composition.}
Replacing random composition with area ordering further raises Stitch Acc.\ from $0.698$ to $0.712$ and reduces L-HFRD from $4.412$ to $4.163$, while several other metrics slightly worsen.
 Its main benefit is preserving the assigned stitch textures in smaller regions: the highlighted bee and fish regions in \Cref{fig:ablation} illustrate how compositing smaller regions after larger ones helps retain these details where regions overlap or nest.

\subsection{Human Expert Evaluation}
\label{sec:user_study}

\begin{figure}[t]
    \centering
    \includegraphics[width=\columnwidth]{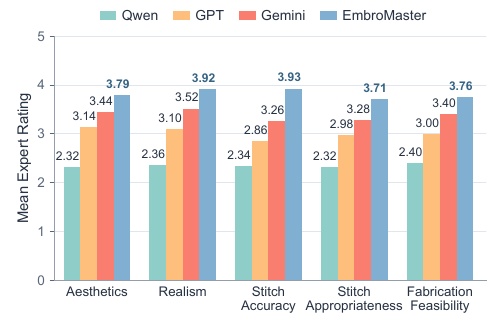}
    \caption{\textbf{Human expert evaluation.}
    Mean ratings from Su embroidery experts across five criteria on a five-point Likert scale.
    }
    \label{fig:user_study}
\end{figure}

Ten Su embroidery experts each evaluate 30 randomly selected comparison groups.
Each group presents the input design, requested style and stitch budget, the stitch assignments and images generated by \methodname and one baseline, with model names hidden and result order randomized.
Experts rate each candidate separately on a five-point Likert scale for \textbf{Aesthetics}, \textbf{Realism}, \textbf{Stitch Accuracy}, \textbf{Stitch Appropriateness}, and \textbf{Fabrication Feasibility} (suitability as a reference for physical embroidery).
Scores are averaged by method and criterion.
\Cref{fig:user_study} shows that \methodname receives the highest mean ratings on all five criteria.

\subsection{Applications}
\label{sec:applications}
\begin{figure}[t]
    \centering
    \includegraphics[width=\columnwidth]{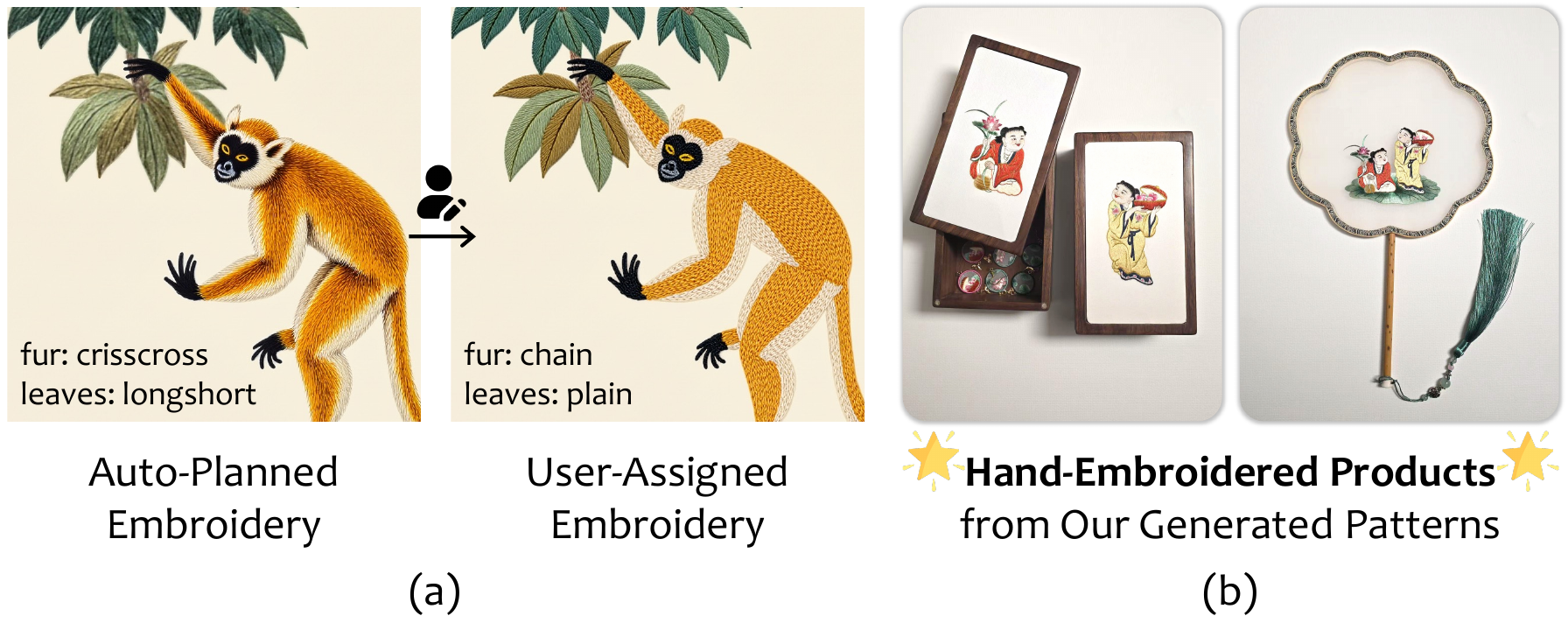}
    \caption{\textbf{Applications.} (a) Users customize stitch assignments. (b) A Su embroidery expert hand-embroiders physical products from patterns generated by \methodname.}
    \label{fig:applications}
\end{figure}

We showcase two representative applications of \methodname in \Cref{fig:applications}. 
First, EmbroMaster enables users to interactively assign stitch types to selected regions, supporting design customization and previews of stitch combinations. Second, its generated patterns can serve as references for physical embroidery. We demonstrate this application by inviting a Su embroidery expert to hand-embroider physical products using these patterns, highlighting the value of our method for exhibitions and cultural promotion.

\section{Limitations and Discussion}
\label{sec:limitations_discussion}

\begin{figure}[t]
    \centering
    \includegraphics[trim=205 255 240 210, clip, width=\columnwidth]{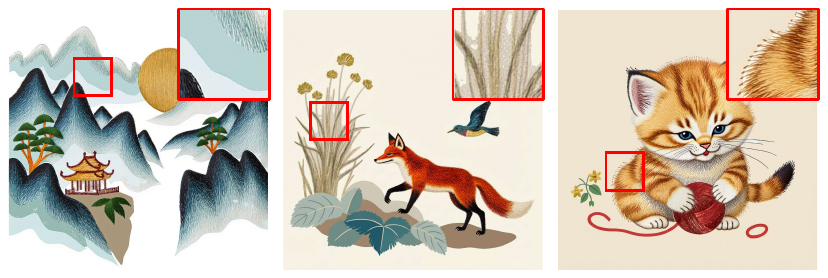}
    \caption{\textbf{Failure cases.} These failures arise from incomplete or inaccurate region masks. Red boxes highlight the affected areas.}

    \label{fig:failure_cases}
\end{figure}

\methodname has two main limitations.
Its spatial fidelity depends on SAM 3 masks: incomplete coverage can leave regions without their assigned stitch, while inaccurate boundaries can cause visible seams, as shown in \Cref{fig:failure_cases}.
In addition, rendering a full image for every required stitch requires several generation calls, which limits interactive efficiency.
To address these limitations, future work could explore a unified model with region-level stitch control to reduce overhead and reliance on mask-based composition while preserving stitch identities and design structure.

\section{Conclusion}
We presented \methodname, a framework that decomposes controllable Su embroidery generation into craft-aware planning and multi-stitch synthesis.
\datasetname combines real and synthetic design--stitch pairs with stitch craft knowledge.
A craft-aware planning agent produces an editable embroidery blueprint that guides stitch-specific rendering and area-ordered composition, supporting design preservation and region-level stitch editing.
Comparative experiments, ablations, and expert evaluation demonstrate the framework's effectiveness in preserving input designs, generating high-quality stitch textures consistent with stitch assignments, and adhering to craft rules.
By making craft decisions explicit and editable, \methodname moves beyond appearance imitation toward an interpretable workflow for embroidery design and interactive exploration.
We will release our code and data for future research.
\section*{LLM Usage Statement}
In this work, a generative large language model (OpenAI GPT) was used to help polish the language of our manuscript, search for related research, and help with code writing and checking.
All research ideas and experimental designs were developed independently by the authors, who also created all figures and tables. The authors have reviewed all content of the submission and take full responsibility for the overall quality of the entire manuscript.

{
    \small
    \bibliographystyle{ieeenat_fullname}
    \bibliography{main}
}










\clearpage
\setcounter{page}{1}
\maketitlesupplementary
\appendix

\etocsettagdepth{main}{none}
\etocsettagdepth{supplementary}{subsection}

\renewcommand{\contentsname}{Contents}
{\small
\tableofcontents
}


\etocdepthtag.toc{supplementary}







\section{\datasetname Dataset}
\label{sec:supp_SuStitch}
\datasetname comprises design--stitch pairs for stitch-specific rendering and stitch craft knowledge for craft-aware planning.
This section details data sources and statistics, stitch craft knowledge, and paired design--stitch data curation.

\subsection{Data Sources and Statistics}
\label{sec:supp_data_statistics}

\begin{figure*}[h]
    \centering
    \includegraphics[trim=10 68 10 70, clip, width=\textwidth]{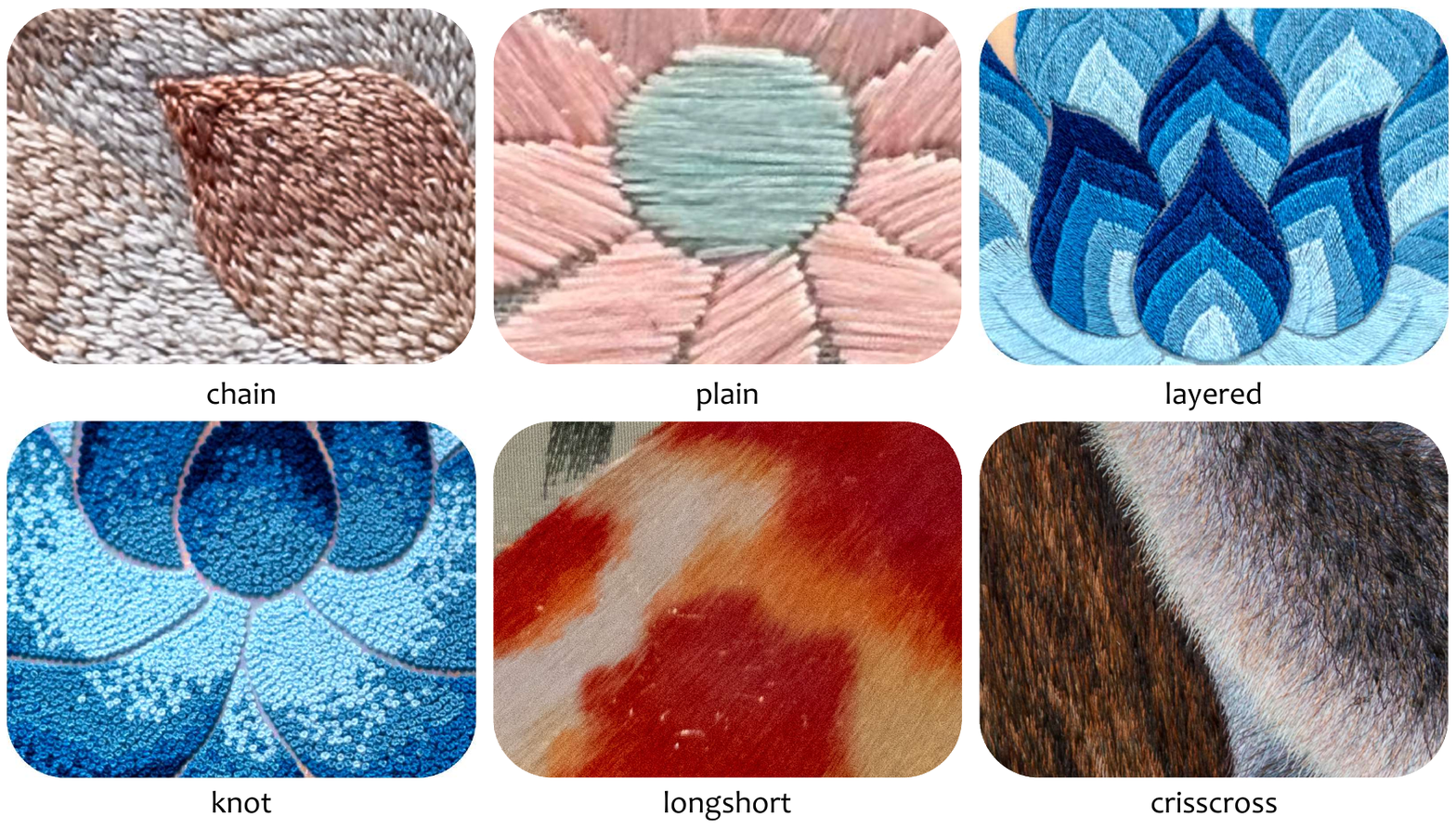}
    \caption{\textbf{Real Su embroidery samples for the six stitch types in \datasetname.}
    These examples reveal linked loops, parallel thread arrangements, stratified coverage, raised knots, smooth tonal blending, and intersecting thread directions, respectively.
    }
    \label{fig:realdata}
\end{figure*}

\paragraph{Real Sample Collection.}
We photograph real Su embroidery using a professional high-definition camera at the China National Silk Museum, Suzhou Museum West, Suzhou Silk Museum, China Grand Canal Museum, and China Embroidery Art Museum.
We also collect real samples from individual embroidery artists' galleries in Zhenhu, including those of Yao Jianping, Zou Yingzi, and Xue Jindi.
Isolated stitch examples and works rendered globally with a single stitch are often displayed to introduce visitors to the characteristics of individual techniques.
Such examples are scarce, whereas complete embroidery artworks commonly combine multiple stitch types.
The availability of clean stitch-specific references therefore varies substantially across techniques.
\Cref{fig:realdata} presents close-up photographs of all six representative stitch types, showing the thread textures that guide stitch identification and reference selection.
These real samples provide the visual basis for stitch-specific expansion and paired data curation.

\paragraph{Data Statistics.}
\Cref{tab:supp_data_statistics} reports the quantities for the six stitches used in this paper.
For knot, chain, plain, and layered stitches, the real collection contains fewer than five distinct motifs for each stitch type.
For stitch-specific expansion, we augment the real images for each of these stitches through cropping, flipping, and rotation to obtain 20 training images for its text-to-image LoRA.
In contrast, long-short and crisscross stitches have 44 and 32 collected images, respectively, allowing us to bypass stitch-specific expansion for them.
Paired design--stitch data curation retains 20 high-quality design--stitch pairs per stitch type.
The augmented images are derived from the original motifs and do not represent additional independently collected designs.
Likewise, the 20 retained pairs for each stitch are selected from the candidate pool after reverse conversion; they should not be interpreted as the total number of synthetic candidates produced through stitch-specific expansion.

\begin{table}
    \centering
    \caption{\textbf{Data quantities at each construction stage.}
    Real counts denote motifs for the first four stitches and images for long-short and crisscross.
    Expansion training images are augmented real images used to train the text-to-image LoRAs, excluding generated candidates; retained pairs are the design--stitch pairs selected after conversion and expert filtering.}
    \label{tab:supp_data_statistics}
    \small
    \setlength{\tabcolsep}{3pt}
    \renewcommand{\arraystretch}{1.15}
    \begin{tabular*}{\columnwidth}{@{\extracolsep{\fill}}lccc@{}}
        \toprule
        Stitch & \makecell{Real\\Collection} & \makecell{Expansion\\Training\\Images} & \makecell{Retained\\Design--Stitch\\Pairs} \\
        \midrule
        Knot       & 3 motifs  & 20      & 20 \\
        Chain      & 2 motifs  & 20      & 20 \\
        Plain      & 3 motifs  & 20      & 20 \\
        Layered    & 3 motifs  & 20      & 20 \\
        Long-short & 44 images & Skipped & 20 \\
        Crisscross & 32 images & Skipped & 20 \\
        \bottomrule
    \end{tabular*}
\end{table}

\paragraph{Resolution and Subject Coverage.}
The collected images have heterogeneous high resolutions, including $2384\times4240$, $5168\times3448$, and $5422\times3586$ pixels.
All images used for LoRA training are preprocessed to $1024\times1024$ pixels.
The data cover animals, plants, landscapes, human figures, objects, and compositions combining these subjects.

\subsection{Stitch Craft Knowledge Collection}
\label{sec:supp_stitch_rules}

We collect knowledge from Su embroidery experts and formalize it into stitch craft rules covering stitch taxonomy and characteristics, style level and stitch budget, and special and general rules for stitch assignment.

\paragraph{Stitch Taxonomy and Characteristics.}
The expert vocabulary organizes Su embroidery techniques into three categories according to their characteristic visual structures.
\textit{Point-based} techniques include knot stitch and beading embroidery.
\textit{Line-based} techniques include chain, connecting, rolling, interlocking chain, couching, and gold couching stitches.
\textit{Surface-based} techniques include plain, layered, long-short, crisscross, interlocking, and admixture stitches, as well as net embroidery, pattern picking, braiding embroidery, partly count embroidery, and needle-looping.
Our experiments use six representative techniques spanning these categories: knot, chain, plain, layered, long-short, and crisscross.
Their visual characteristics and typical roles are summarized in \Cref{tab:supp_stitch_vocabulary}.
We use \texttt{longshort} as the identifier for long-short stitch in prompts and model names.

\begin{table*}[t]
    \centering
    \caption{\textbf{The six representative stitches used in \methodname.}
    }
    \label{tab:supp_stitch_vocabulary}
    \small
    \setlength{\tabcolsep}{5pt}
    \renewcommand{\arraystretch}{1.15}
    \begin{tabular}{@{}p{\dimexpr(\textwidth-6\tabcolsep)*11/100\relax}p{\dimexpr(\textwidth-6\tabcolsep)*9/100\relax}p{\dimexpr(\textwidth-6\tabcolsep)*40/100\relax}p{\dimexpr(\textwidth-6\tabcolsep)*40/100\relax}@{}}
        \toprule
        {\raggedright Stitch\par} & {\raggedright Category\par} & {\raggedright Characteristic morphology\par} & {\raggedright Typical role\par} \\
        \midrule
        {\raggedright Knot\par} & {\raggedright Point\par} & {\raggedright Dense, bead-like knots with a raised granular appearance.\par} & {\raggedright Dotted details and granular elements, particularly flower stamens.\par} \\
        {\raggedright Chain\par} & {\raggedright Line\par} & {\raggedright Continuous linked loops with strong structural lines and color transitions along the stitching direction.\par} & {\raggedright Linear structures and decorative outlines; often used independently.\par} \\
        {\raggedright Plain\par} & {\raggedright Surface\par} & {\raggedright Parallel, even, non-overlapping threads whose endpoints follow the motif boundaries.\par} & {\raggedright Small petals, leaves, stems, facial details, and local clothing details.\par} \\
        {\raggedright Layered\par} & {\raggedright Surface\par} & {\raggedright Successive batches of plain stitches connect across layers, yielding orderly stratification and gradual color transitions.\par} & {\raggedright Larger regions that require layered coverage and organized tonal transitions.\par} \\
        {\raggedright Long-short\par} & {\raggedright Surface\par} & {\raggedright Overlapping stitches of varying lengths form a smooth surface with natural color blending.\par} & {\raggedright Delicate tonal modeling and smooth, realistic transitions.\par} \\
        {\raggedright Crisscross\par} & {\raggedright Surface\par} & {\raggedright Lines and layers intersect at small angles, producing a volumetric, fluffy appearance with smooth tonal transitions.\par} & {\raggedright Directional, multilayered textures such as animal fur.\par} \\
        \bottomrule
    \end{tabular}
\end{table*}

\begin{table*}[t]
    \centering
    \caption{\textbf{General rules for stitch assignment.}
    }
    \label{tab:supp_semantic_rules}
    \small
    \setlength{\tabcolsep}{4pt}
    \renewcommand{\arraystretch}{1.14}
    \begin{tabular}{@{}p{\dimexpr(\textwidth-8\tabcolsep)*11/100\relax}p{\dimexpr(\textwidth-8\tabcolsep)*17/100\relax}p{\dimexpr(\textwidth-8\tabcolsep)*24/100\relax}p{\dimexpr(\textwidth-8\tabcolsep)*24/100\relax}p{\dimexpr(\textwidth-8\tabcolsep)*24/100\relax}@{}}
        \toprule
        {\raggedright Category\par} & {\raggedright Semantic region\par} & {\raggedright Decorative style\par} & {\raggedright Transitional style\par} & {\raggedright Realistic style\par} \\
        \midrule
        {\raggedright Plants\par} & {\raggedright Petals\par} & {\raggedright Knot, plain, layered\par} & {\raggedright Layered, plain, long-short\par} & {\raggedright Long-short, crisscross\par} \\
        & {\raggedright Leaves\par} & {\raggedright Knot, plain, layered\par} & {\raggedright Plain, layered, long-short\par} & {\raggedright Long-short, crisscross\par} \\
        & {\raggedright Fruits\par} & {\raggedright Knot, plain, layered\par} & {\raggedright Layered, long-short\par} & {\raggedright Long-short, crisscross\par} \\
        & {\raggedright Branches/stems\par} & {\raggedright Plain, layered\par} & {\raggedright Long-short\par} & {\raggedright Long-short, crisscross\par} \\
        \midrule
        {\raggedright Animals\par} & {\raggedright Body\par} & {\raggedright Knot\par} & {\raggedright Long-short\par} & {\raggedright Long-short, crisscross\par} \\
        & {\raggedright Opaque wings\par} & {\raggedright Knot, layered\par} & {\raggedright Layered, long-short\par} & {\raggedright Long-short, crisscross\par} \\
        \midrule
        {\raggedright Figures\par} & {\raggedright Face\par} & {\raggedright Plain\par} & {\raggedright Long-short\par} & {\raggedright Long-short, crisscross\par} \\
        & {\raggedright Clothing\par} & {\raggedright Knot\par} & {\raggedright Plain, long-short\par} & {\raggedright Long-short, crisscross\par} \\
        \midrule
        {\raggedright Landscapes\par} & {\raggedright Mountains/rivers\par} & {\raggedright Knot, layered\par} & {\raggedright Long-short\par} & {\raggedright Long-short, crisscross\par} \\
        & {\raggedright Buildings\par} & {\raggedright Knot\par} & {\raggedright Plain, layered, long-short\par} & {\raggedright Long-short, crisscross\par} \\
        \midrule
        {\raggedright Objects\par} & {\raggedright General items\par} & {\raggedright Knot, layered\par} & {\raggedright Long-short\par} & {\raggedright Long-short, crisscross\par} \\
        \bottomrule
    \end{tabular}
\end{table*}

\paragraph{Style Level and Stitch Budget.}
The rule base describes three styles: decorative, transitional, and realistic.
The experiments use only the decorative and realistic styles; the transitional style is included here to document the complete stitch craft knowledge.
\textit{Decorative style} emphasizes abstraction, contrast, and orderly patterns, with chain, plain, layered, and knot as its default stitch palette. Chain stitch is used independently and should not be combined with other stitch types.
\textit{Transitional style} balances orderly patterns with natural blending, using plain, layered, knot, and long-short stitches, either independently or in combination.
\textit{Realistic style} emphasizes natural appearance, smooth tonal transitions, and layered thread structures, primarily using long-short and crisscross stitches.
Knot and plain stitches are reserved for appropriate local details rather than used alone over an entire realistic design.
The object-specific rules below provide exceptions where supported by the visible content.
The stitch budget specifies the exact number of distinct stitch types used across all named objects, rather than the number of semantic regions.

\paragraph{Special Rules (High-Priority Empirical Constraints).}
The expert rule base further specifies four constraints that take precedence over general semantic suggestions:
\begin{enumerate}
    \item \textbf{Flower stamens:} use knot stitch to express their raised granular structure.
    \item \textbf{Animal fur:} use long-short or crisscross stitch to express directional thread textures.
    \item \textbf{Transparent wings:} normally use long-short stitch; realistic styles may also use crisscross stitch.
    \item \textbf{Region scale for plain stitch:} reserve plain stitch for small elements, such as small petals, leaves, stems, facial details, and local clothing details. For a large region requiring stratified coverage, use layered stitch, which applies plain stitches in successive connected batches.
\end{enumerate}

\paragraph{General Rules for Stitch Assignment.}
\Cref{tab:supp_semantic_rules} lists semantic candidates for decorative, transitional, and realistic embroidery styles.
The craft-aware planning agent applies the special rules before the general semantic stitch candidates while respecting the exact stitch budget and assigning one stitch to each named object.
When a preferred assignment conflicts with these hard requirements, it selects the closest professionally appropriate alternative. The prompt in \Cref{fig:prompt2} specifies how these rules are applied.

\subsection{Paired Design-Stitch Data Curation}
\label{sec:supp_bootstrapping}

\begin{figure*}[t]
    \centering
    \includegraphics[trim=33 130 25 88 , clip, width=\textwidth]{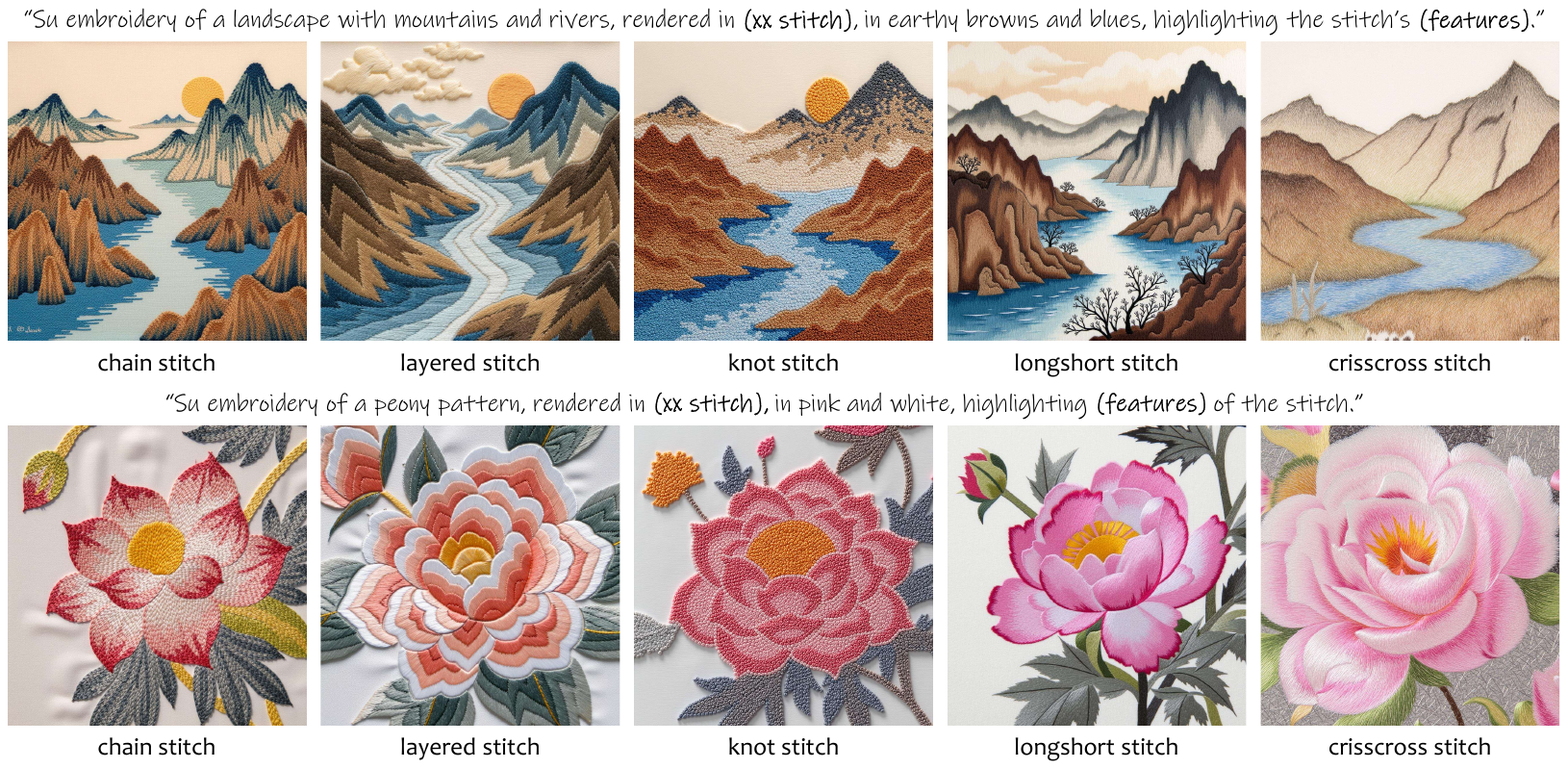}
    \caption{\textbf{Text-to-image synthesis with stitch-specific FLUX.1-dev LoRAs.}
    Within each row, the subject description and latent seed are held fixed while the target stitch and its characteristic features vary.
    }
    \label{fig:t2i}
\end{figure*}

We construct design--stitch pairs through stitch-specific expansion, embroidery-to-design conversion, and expert filtering.
For stitch-specific expansion, we augment the real images of each sparsely represented stitch through cropping, flipping, and rotation to obtain the 20-image training set reported in \Cref{tab:supp_data_statistics}.
Each training image is captioned with its subject, stitch identity, and characteristic using the template
\begin{quote}
    \small\normalfont\itshape\raggedright
    ``Su embroidery of (object), rendered in (stitch), highlighting (features).''
\end{quote}
Here, \textit{(object)} describes the depicted subject, \textit{(stitch)} names the target technique, and \textit{(features)} describes its visual characteristics from the expert vocabulary.
We fine-tune a separate LoRA~\cite{hu2022lora} on FLUX.1-dev~\cite{blackforestlabs2024flux1} for each stitch, with rank 32, alpha 32, and a learning rate of $1\times10^{-4}$.
The learned adapter generates single-stitch embroidery candidates covering diverse subjects using the same prompt template, with the subject varied while the stitch identity and characteristic remain specified.
This expands the candidate pool beyond the few collected motifs, although the candidates vary in quality and in the clarity of their stitch characteristics.
Long-short and crisscross bypass stitch-specific expansion because sufficiently many real images are available.
For the visualization in \Cref{fig:t2i}, we additionally train text-to-image LoRAs for these two stitches solely to demonstrate their learned appearances; these adapters are not used for data construction.
During paired design--stitch data curation, Nano Banana Pro~\cite{google2025nanobananapro} converts real and synthetic embroideries into flat-color designs.
The embroidery-to-design conversion uses the prompt
\begin{quote}
    \small\normalfont\itshape\raggedright
    ``Turn the embroidery artwork into a smooth digital painting with flat colors, retaining the structure and colors, removing all texture.''
\end{quote}
The source embroidery is retained as the target, while the converted flat design serves as the input for subsequent renderer training, avoiding the need for a manually drawn counterpart.
Experts inspect the candidate pairs based on stitch fineness, stitch direction, color harmony, aesthetic sense, and structure alignment, retaining 20 high-quality aligned design--stitch pairs per stitch.
This filtering removes poor generations and unsuitable conversions, yielding pixel-aligned supervision for all six stitch-specific renderers, including those whose data do not require stitch-specific expansion.
This alignment helps each renderer learn the intended stitch appearance while preserving object geometry and spatial correspondence with the input design.

\section{Implementation Details}
\label{sec:supp_implementation}
This section specifies the implementation details and prompting of the compared baseline methods and the training of stitch-specific renderers.
\begin{figure*}[t]
    \centering
    \includegraphics[trim=38 195 32 120, clip, width=\textwidth]{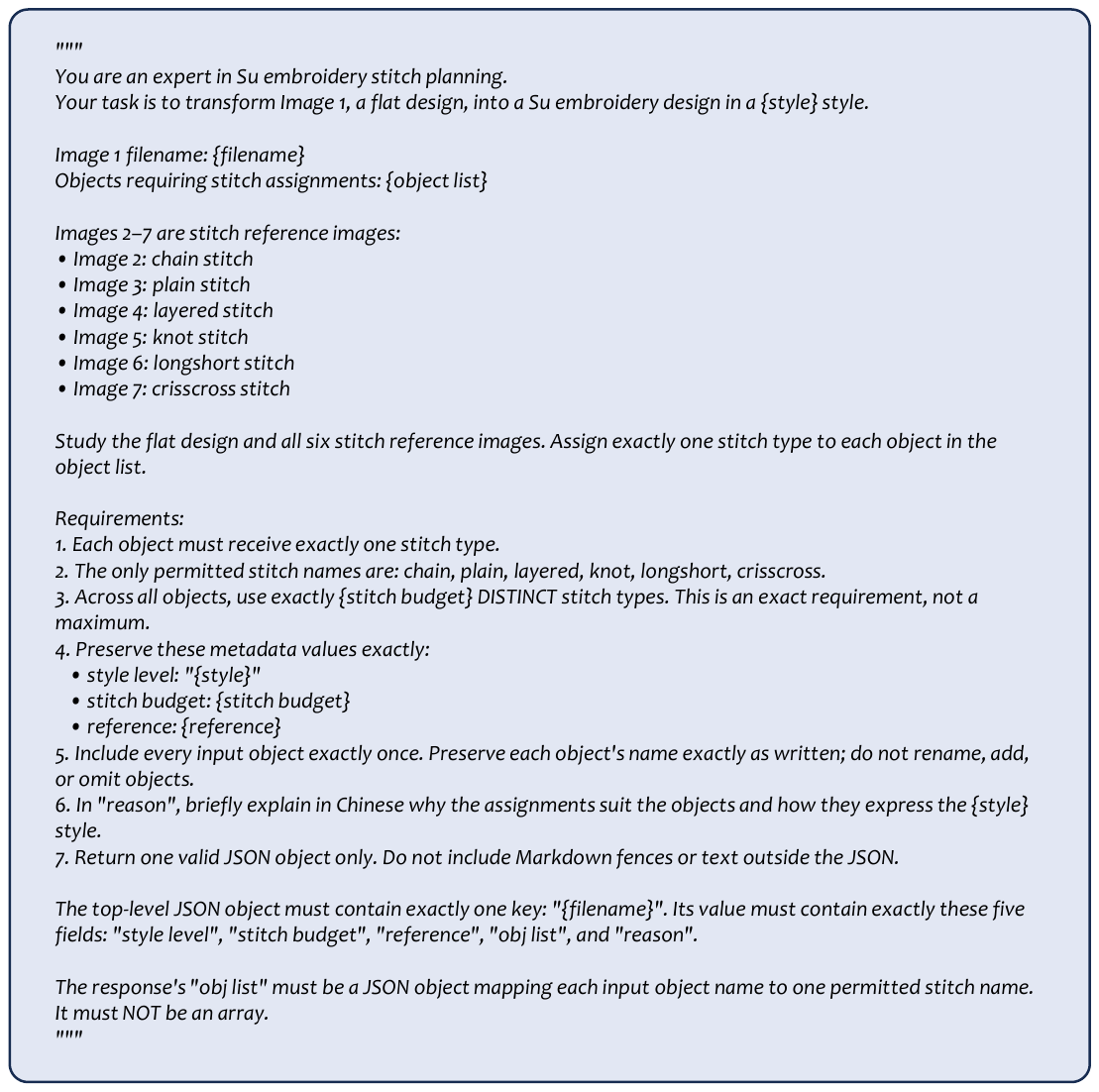}
    \caption{\textbf{Shared stitch-planning prompt.}
    All planners receive the flat design, a fixed object list, six stitch reference images, and the requested style and exact stitch budget.
    The template requires one stitch assignment per object and a structured JSON response. 
    }
    \label{fig:prompt1}
\end{figure*}

\subsection{Implementation Details and Prompt Templates}
\label{sec:supp_implementation_prompts}

\paragraph{Compared Pipelines.}
The general-purpose pipelines combine a multimodal planner with an image generator: Qwen3.8-Flash-Next~\cite{qwen2026qwen38flashnext} with Qwen-Image-Edit-2511~\cite{qwen2025qwenimageedit2511}, GPT-6 Sol~\cite{openai2026gpt6sol} with GPT-Image-2~\cite{openai2026gptimage2}, and Gemini 3.8 Flash~\cite{google2026gemini38flash} with Nano Banana Pro~\cite{google2025nanobananapro}, as specified in the main paper.
\methodname uses GPT-6 Sol as the reasoning backbone of the craft-aware planning agent and SAM 3~\cite{carion2026sam3} for mask generation.
Its multi-stitch synthesis stage follows the editable embroidery blueprint through stitch-specific rendering with FLUX.1-Kontext-dev~\cite{batifol2025fluxkontext} LoRAs and area-ordered composition.
The renderer training configuration is given in \Cref{sec:supp_renderer_training} and the main paper.
For the end-to-end comparisons, each method's own predicted assignments define the target stitch classes for stitch identity and class-conditioned texture evaluation.
For the fixed-assignment experiment in \Cref{sec:supp_fixed_assignments}, all methods instead use the same supplied object-to-stitch mapping.
Rendering fidelity and craft-rule compliance are therefore examined separately.
For each end-to-end test case, all methods receive the same flat design, object list, six stitch reference images, style level, and stitch budget.
The planning input contains seven images in a fixed order: the flat design first, followed by references for chain, plain, layered, knot, long-short, and crisscross stitches.
The style level and stitch budget are identical across methods for the same test case.
The stitch craft knowledge block is supplied only to \methodname as a component of craft-aware planning; the general-purpose baselines receive the shared planning prompt without this additional knowledge.


\paragraph{Shared Stitch-Planning Prompt.}
\Cref{fig:prompt1} presents the shared planning template.
It fixes the permitted stitch identifiers, preserves all input object names, and requires exactly the requested number of distinct stitch types.
Placeholders in braces are replaced with the metadata of the current test case.
The output contains the style, stitch budget, reference metadata, object-to-stitch mapping, and a brief rationale.

\begin{figure*}[t]
    \centering
    \includegraphics[trim=6 100 6 100, clip, width=\textwidth]{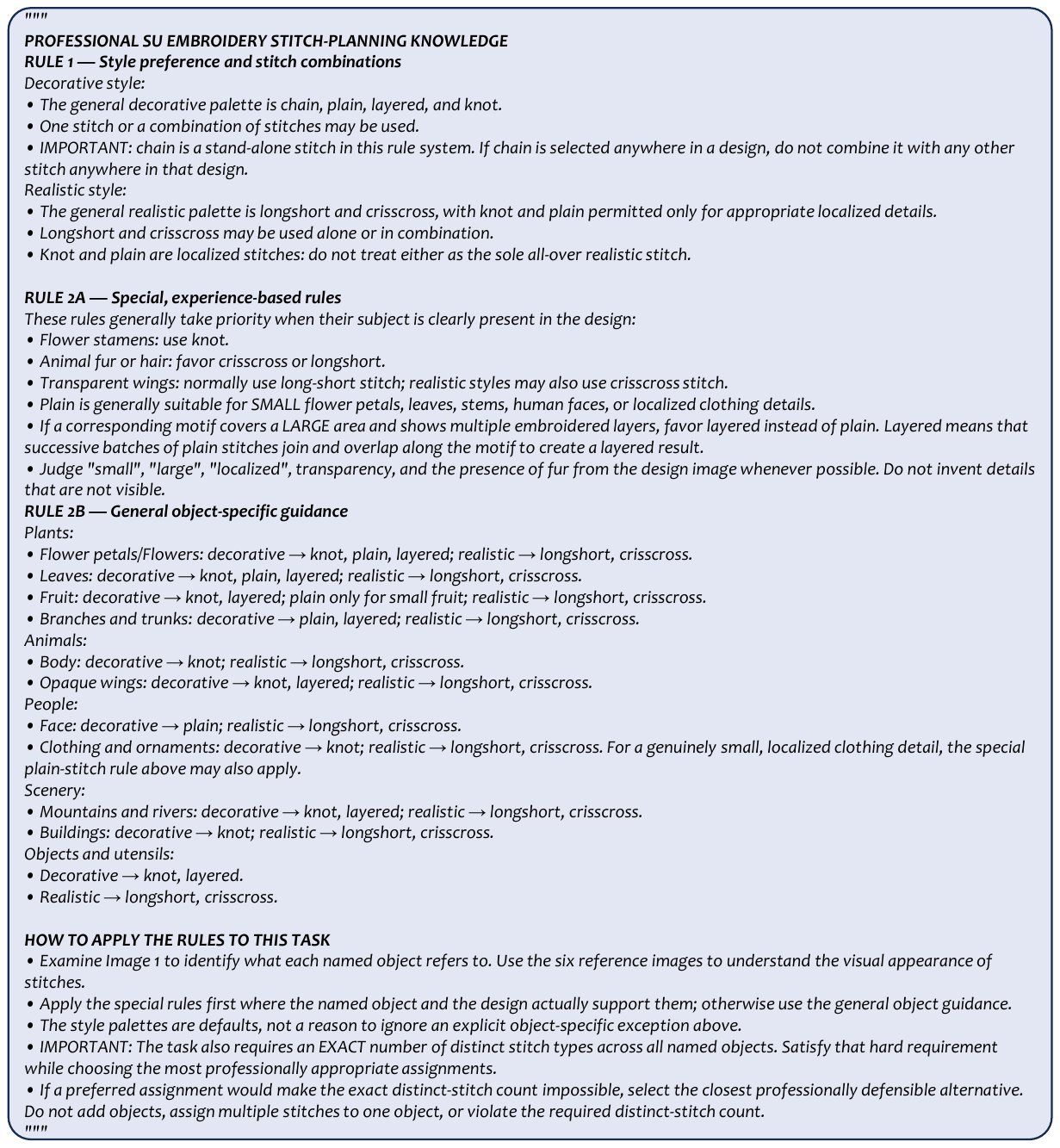}
    \caption{\textbf{Stitch craft knowledge supplied to \methodname.}
    }
    \label{fig:prompt2}
\end{figure*}

\paragraph{Stitch Craft Knowledge Prompt for \methodname.}
For \methodname, we augment the shared planning template with the stitch craft knowledge block in \Cref{fig:prompt2}, which operationalizes the rules summarized in \Cref{sec:supp_stitch_rules}.
The special rules take precedence over general semantic suggestions when supported by the visible content.
The exact stitch budget and one-stitch-per-object requirements remain hard constraints throughout the entire planning process.

\begin{figure*}[t]
    \centering
    \includegraphics[trim=25 480 22 150, clip, width=\textwidth]{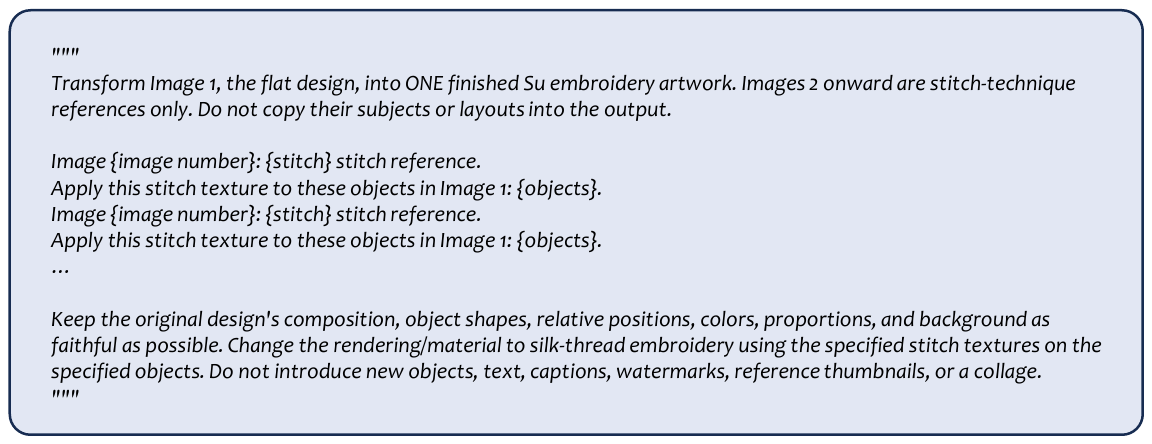}
    \caption{\textbf{Multi-stitch synthesis prompt for the baseline image generators.}
    The template associates each stitch reference with the objects assigned to that stitch.
    It instructs the generator to preserve the flat design's content, geometry, and colors while changing its material appearance to embroidery.
    Reference images specify stitch texture rather than output subjects or layouts.}
    \label{fig:prompt3}
\end{figure*}

\paragraph{Multi-Stitch Synthesis Prompt for Baseline Methods.}
After planning, each baseline image generator receives the flat design, stitch references, and object-to-stitch assignments through the template in \Cref{fig:prompt3}.
The reference block is repeated for the supplied stitch references; the image-number placeholder identifies a reference's position in the generation input, and \texttt{\{objects\}} lists the objects assigned to that stitch.
The prompt separates the texture examples from the design content to preserve.
For the fixed-assignment comparison in \Cref{sec:supp_fixed_assignments}, the same template is populated with the common assignments instead of a baseline's own predicted plan.
\methodname combines the object-level stitch assignments with SAM 3 masks to form the editable embroidery blueprint.
Each required stitch-specific renderer produces one global single-stitch image from the flat design, using the instruction in \Cref{sec:supp_renderer_training}.
The blueprint masks select the assigned regions, which are combined through area-ordered composition.

\subsection{Stitch-Specific Renderer Training}
\label{sec:supp_renderer_training}

\begin{figure*}
    \centering
    \includegraphics[trim=29 140 28 110 , clip, width=\textwidth]{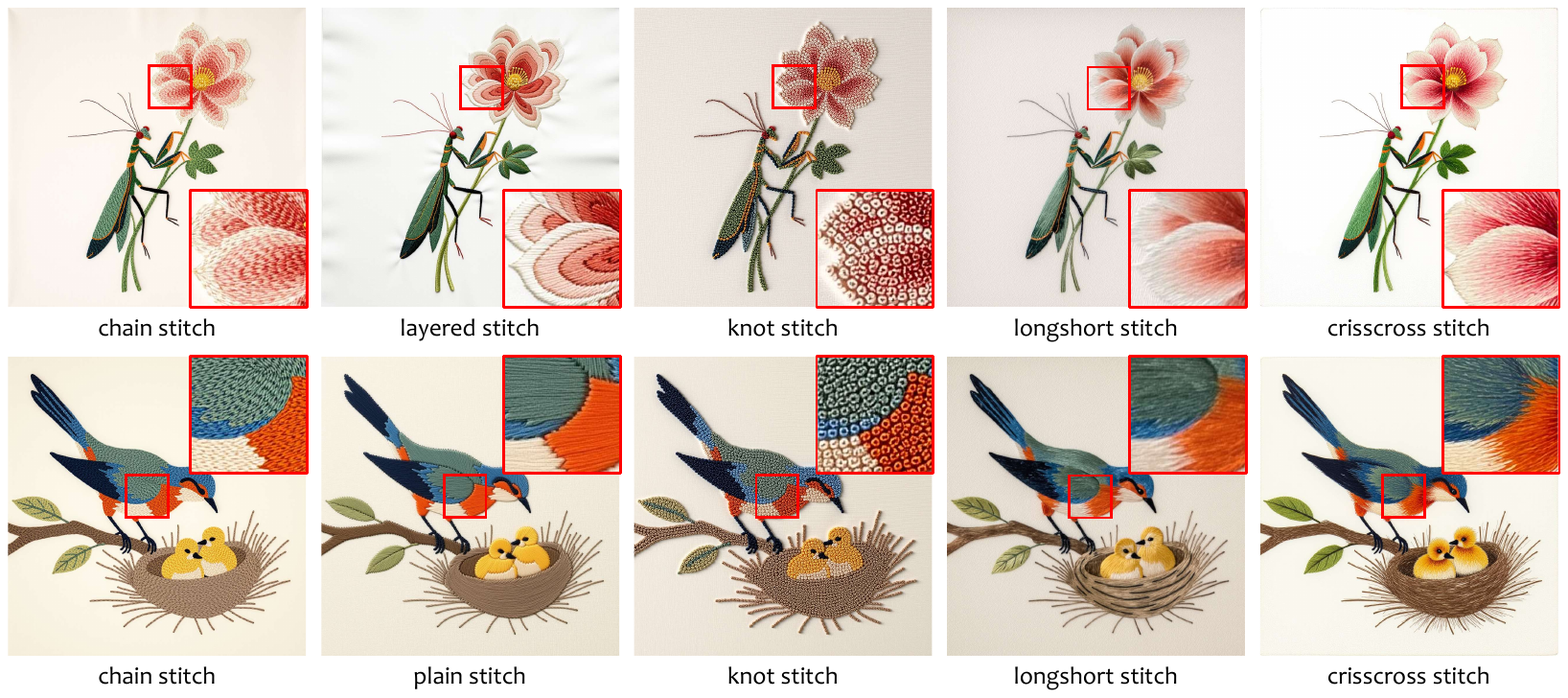}
    \caption{\textbf{Image-to-image synthesis with stitch-specific FLUX.1-Kontext-dev LoRAs.}
    Each row shows global stitch renderings of the same flat design.
    Red boxes and enlarged insets highlight the local thread structures.}
    \label{fig:i2i}
\end{figure*}
For each of the six stitches, we train an independent stitch-specific renderer by LoRA fine-tuning FLUX.1-Kontext-dev on its 20 curated design--stitch pairs.
Each pair is associated with the instruction
\begin{quote}
    \small\normalfont\itshape\raggedright
    ``convert to xxx\_stitch embroidery style''
\end{quote}
where \texttt{xxx} is replaced by the stitch identifier: \texttt{knot}, \texttt{chain}, \texttt{plain}, \texttt{layered}, \texttt{longshort}, or \texttt{crisscross}.
The corresponding \texttt{xxx\_stitch} token serves as the trigger word.
We use rank 32, alpha 32, and a learning rate of $1\times10^{-4}$ for each renderer.
These adapters learn to transform an input design into its stitch-specific rendition while preserving the design structure.
As shown in \Cref{fig:i2i}, the renderers produce distinct stitch appearances for the mantis-and-flower and bird-and-nest designs; the enlarged regions reveal differences in thread arrangement and stitch textures.
The FLUX.1-dev adapters support stitch-specific expansion during dataset construction, whereas the FLUX.1-Kontext-dev adapters implement stitch-specific rendering in the multi-stitch synthesis stage.

\section{Evaluation Details}

\label{sec:supp_evaluation}

This section specifies the evaluation datasets, metric definitions and computation protocols, and user study details.

\subsection{Evaluation Datasets}
\label{sec:supp_evaluation_data}

We generate and select 200 flat designs containing multiple objects as the evaluation dataset.
Before running any compared pipeline, we use Qwen3-VL-8B-Instruct~\cite{bai2025qwen3vl} to identify the objects in each design and fix the resulting object list for all methods.
Thus, object discovery is performed once in a shared preprocessing step, allowing the comparison to focus on stitch assignment and multi-stitch synthesis.

For the end-to-end comparisons in the main paper, we use an exact stitch budget of $b\in\{2,3\}$.
Every named object receives one stitch assignment, and the complete plan must contain exactly $b$ distinct stitch types.
Although \methodname supports more than three stitch types in a design, the general-purpose baselines already exhibit planning and rendering errors at the two- and three-stitch settings.
We therefore use this shared range to compare the methods at a controlled compositional complexity, rather than increasing the required stitch count beyond the settings considered in the common benchmark.
The number of objects is independent of $b$: multiple objects may share the same stitch.
The fixed-assignment rendering experiment in \Cref{sec:supp_fixed_assignments} uses the same 200 designs with common object-to-stitch assignments across methods.
The additional results in \Cref{sec:supp_results} also illustrate user-edited designs with four to six stitch types, beyond stitch budgets used in the main paper.

\subsection{Metric Definitions and Computation}
\label{sec:supp_metric_protocols}

Let $D_i$ and $I_i$ denote the design and generated image of test case $i$, and let $N$ be the number of evaluated input--output pairs.
The evaluation expects $1024\times1024$ RGB images and a shared set of object masks associated with the input designs.
For the spatial and color metrics, masks are binarized at intensity $127$, and regions containing fewer than 32 foreground pixels are excluded.
We denote the eligible image--region pairs by $\mathcal{R}$, their binary masks by $M_r$, and their foreground areas by $|M_r|$.
Pixel means are computed within each mask before taking an equal-weight mean across regions; region area is not used as an aggregation weight.
All formulas below define raw, unscaled metric outputs.
For display in the main paper's quantitative comparison and ablation tables and in \Cref{tab:qualitative_suppl}, LPIPS, RSP, Histogram Loss, and Local KID are multiplied by $100$, and Local HFRD is multiplied by $1000$.
Local Color $\Delta E_{00}$, Stitch Classification Accuracy, and Stitch Assignment Rule Compliance are reported without scaling.
These display factors do not change the metric definitions.

\paragraph{LPIPS ($\downarrow$).}
We measure full-image perceptual discrepancy using AlexNet LPIPS~\cite{zhang2018lpips}.
Let $T_{256}$ bilinearly resize an image to $256\times256$ pixels and normalize its RGB values to $[-1,1]$.
The dataset score is
\begin{equation}
    \mathrm{LPIPS}
    =\frac{1}{N}\sum_{i=1}^{N}
    d_{\mathrm{Alex}}\!\left(T_{256}(D_i),T_{256}(I_i)\right).
    \label{eq:supp_lpips}
\end{equation}
No foreground mask is applied.
This design-to-output comparison evaluates perceptual content preservation.

\paragraph{Region-wise Structure Preservation (RSP, $\uparrow$).}
We compute a foreground-only, low-pass SSIM score~\cite{wang2004ssim} to reduce sensitivity to thread-scale texture.
For each mask, we extract its bounding box with 11 pixels of context, clipped to the image boundary.
The two images are converted to grayscale in $[0,1]$.
In each crop, pixels outside the mask are filled with that image's foreground mean before Gaussian filtering with $\sigma=1.5$ pixels.
After filtering, the outside-mask values are again replaced by the filtered foreground mean.
For a grayscale crop $x$, this preprocessing can be written as
\begin{equation}
\begin{aligned}
    \mu_r(x)&=\frac{1}{|M_r|}\sum_u M_r(u)x(u),\\
    F_r(x)&=M_r\odot x+(1-M_r)\mu_r(x),\\
    P_r(x)&=F_r\!\left(G_{1.5}*F_r(x)\right),
\end{aligned}
\label{eq:supp_structure_preprocess}
\end{equation}
where $G_{1.5}$ is the Gaussian kernel and $u$ indexes crop pixels.
We obtain a full SSIM map $S_r$ from the two preprocessed crops using Gaussian window weights, window size up to $11\times11$, window standard deviation $1.5$, and data range $1$.
The window is reduced to an odd size for smaller crops; other SSIM parameters use the scikit-image defaults.
Only foreground values of the map enter the score:
\begin{equation}
    \mathrm{RSP}
    =\frac{1}{|\mathcal{R}|}
    \sum_{r\in\mathcal{R}}
    \frac{\sum_u M_r(u)S_r(u)}{|M_r|}.
    \label{eq:supp_rsp}
\end{equation}
The outer average is over all valid regions, rather than an average of image-wise region means.

\paragraph{Histogram Loss (Hist. Loss, $\downarrow$).}
Prior work has explored differentiable histogram-based objectives for image-to-image translation~\cite{aviaharon2023histogram}.
In this work, we measure the discrepancy between the global color distributions of the design and generated image using the Hellinger distance between their normalized joint RGB histograms. For each complete image, we compute a normalized joint RGB histogram with 8 bins per channel, yielding $8^3=512$ bins. Let $p_i$ and $q_i$ denote the design and generated-image histograms. The reported discrepancy is
\begin{equation}
    L_{\mathrm{hist}}
    =\frac{1}{N}\sum_{i=1}^{N}
    \sqrt{1-\sum_{k=1}^{512}\sqrt{p_{ik}q_{ik}}}.
    \label{eq:supp_histogram}
\end{equation}
This is the Hellinger distance between the normalized joint RGB histograms.
It includes the background and does not use region masks.

\paragraph{Local Color $\Delta E_{00}$ (L-$\Delta E_{00}$, $\downarrow$).}
Both images are converted to CIELAB using the D65 illuminant and the $2^\circ$ standard observer.
We compute the standard pixelwise CIEDE2000 color difference~\cite{luo2001ciede2000}, then average within each mask and across regions:
\begin{equation}
\begin{aligned}
    L_{\Delta E_{00}}
    &=\frac{1}{|\mathcal{R}|}\sum_{r\in\mathcal{R}}\delta_r,\\
    \delta_r
    &=\frac{1}{|M_r|}\sum_u M_r(u)\,
    \Delta E_{00}\!\left(D_r^{\mathrm{Lab}}(u),I_r^{\mathrm{Lab}}(u)\right).
\end{aligned}
\label{eq:supp_deltae}
\end{equation}
Thus, a large region contributes the same outer weight as a small eligible region.

\paragraph{Atlas Construction for Local Texture Evaluation.}
For each target stitch class (from the method's own plan in the end-to-end comparisons, or from the common plan in the fixed-assignment comparison), we construct 50 mask-blended $1024\times1024$ atlases from its generated object regions.
Atlas layouts are determined from masks and stitch assignments, with random seed 42, without inspecting the generated RGB texture values.
Only translation and mask alpha blending are used: regions are not rotated, rescaled, or split into connected components, and original mask overlaps are retained.
The background is white, and mask feathering uses Gaussian $\sigma=1.0$.
The provided configuration uses 24 placement candidates, at least four placements per atlas, and at most three objects from the same source image.
Each atlas must have at least $55\%$ foreground occupancy in its central $512\times512$ crop, which is used at native resolution for texture evaluation.
Real-reference images are also $1024\times1024$ and contribute their central $512\times512$ crops, without foreground masking or reference exclusions.
All available reference images in the corresponding stitch class are used.
The six examples supplied to the planners are visual prompting inputs; the texture metrics use the class-specific reference library.
Atlas sampling can reuse source regions, so the 50 atlases are not 50 independently collected embroidery artworks.
Identical masks and stitch assignments share cached layouts; methods with different plans can have different layouts and assigned class sets.

\paragraph{Local HFRD (L-HFRD, $\downarrow$).}
Following the HFRD metric introduced in EmoLoRA~\cite{ma2025emolora}, we adapt it to local texture evaluation.
For a native center crop $P$, we use OpenCV's 8-bit grayscale conversion and compute its shifted two-dimensional FFT energy $E_P(v)=|\mathcal{F}(P)(v)|^2$.
With frequency radius measured in FFT bins from the center, the high-frequency ratio is
\begin{equation}
    h(P)=\frac{\sum_{\|v\|_2>46}E_P(v)}{\sum_v E_P(v)}.
    \label{eq:supp_hf_ratio}
\end{equation}
The denominator includes the DC component; no mean subtraction or Hann window is applied.
Let $A_{s,a}$ be atlas crop $a$ for stitch $s$, let $\mathcal{T}_s$ be the real-reference crops of that stitch, and let $\mathcal{C}$ be the target stitch classes.
We first compute the reference mean and then average the absolute discrepancy of each atlas:
\begin{equation}
\begin{aligned}
    \bar h_s&=\frac{1}{|\mathcal{T}_s|}\sum_{P\in\mathcal{T}_s}h(P),\\
    d_s&=\frac{1}{50}\sum_{a=1}^{50}\left|h(A_{s,a})-\bar h_s\right|,\\
    \mathrm{L\text{-}HFRD}
    &=\frac{1}{|\mathcal{C}|}\sum_{s\in\mathcal{C}}d_s.
\end{aligned}
\label{eq:supp_hfrd}
\end{equation}
The absolute value is taken before averaging over atlases.
This atlas protocol produces a stitch-class texture statistic, not a per-object HFRD score.

\paragraph{Local KID (L-KID, $\downarrow$).}
We use the same atlas and reference center crops to compare stitch-specific feature distributions using KID~\cite{binkowski2018kid}.
Crops are resized to $299\times299$ with OpenCV, using area interpolation for downsampling and linear interpolation for upsampling, and converted to RGB for feature extraction.
The torch-fidelity Inception network provides 2048-dimensional features.
For feature vectors $x,y\in\mathbb{R}^{2048}$, the kernel is
\begin{equation}
    k(x,y)=\left(\frac{x^{\mathsf T}y}{2048}+1\right)^3.
    \label{eq:supp_kid_kernel}
\end{equation}
For each class and repetition, we randomly select $m=14$ generated and $m=14$ real-reference features without replacement within each subset.
Both class pools must contain at least 14 samples.
For subsets $X=\{x_a\}_{a=1}^{m}$ and $Y=\{y_a\}_{a=1}^{m}$, we compute
\begin{equation}
\begin{aligned}
    \widehat{\mathrm{MMD}}_u^2(X,Y)
    ={}&\frac{\sum_{a\ne b}k(x_a,x_b)}{m(m-1)}\\
       &+\frac{\sum_{a\ne b}k(y_a,y_b)}{m(m-1)}\\
       &-\frac{2}{m^2}\sum_{a=1}^{m}\sum_{b=1}^{m}k(x_a,y_b).
\end{aligned}
\label{eq:supp_mmd}
\end{equation}
Using random seed 42, we repeat the subset evaluation 100 times and macro-average over the assigned classes:
\begin{equation}
\begin{aligned}
    \mathrm{KID}_s&=\frac{1}{100}\sum_{t=1}^{100}
    \widehat{\mathrm{MMD}}_u^2(X_{s,t},Y_{s,t}),\\
    \mathrm{L\text{-}KID}
    &=\frac{1}{|\mathcal{C}|}\sum_{s\in\mathcal{C}}\mathrm{KID}_s.
\end{aligned}
    \label{eq:supp_kid}
\end{equation}
KID is a distribution-level score and is not computed for individual regions.
The unbiased estimator can be negative, and negative values are retained.
For both texture metrics, unassigned stitch classes are excluded from the primary class average.

\paragraph{Stitch Classification Accuracy (Stitch Acc., $\uparrow$).}
We fine-tune Qwen3-VL-8B-Instruct~\cite{bai2025qwen3vl} as a six-class stitch classifier using LoRA rank 32, alpha 64, and learning rate $5\times10^{-5}$, and use the checkpoint at training step 664.
For inference, each object's tight bounding box is extracted from the generated image, pixels outside its binary mask are filled with white, and the crop is centered on the smallest fitting square canvas from $\{256,512,768,1024\}$ pixels.
Objects are not resized.
The classifier receives only the crop and the following fixed instruction; neither the planned label nor the object name enters its prompt:
\begin{quote}
    \small\normalfont\itshape\raggedright
    Identify the embroidery stitch type shown in this image. Choose exactly one of the following labels: chain, crisscross, knot, layered, longshort, plain. Reply with only the label.
\end{quote}
Inference uses deterministic decoding with a maximum of 16 new tokens.
For the set $\mathcal{O}$ of evaluated objects, let $t_o$ be the stitch assigned by that method's own plan, $C_o$ the preprocessed crop, and $f(C_o)$ the predicted class.
The primary score is object-level micro accuracy:
\begin{equation}
    \mathrm{StitchAcc}
    =\frac{1}{|\mathcal{O}|}\sum_{o\in\mathcal{O}}
    \mathbf{1}\!\left[f(C_o)=t_o\right].
    \label{eq:supp_stitch_acc}
\end{equation}
This score measures whether the generated appearance follows the predicted plan, independently of whether the plan is professionally appropriate.

\paragraph{Stitch Assignment Rule Compliance (Rule Comp., $\uparrow$).}
GPT-5.6 Sol~\cite{openai2026gpt56sol} serves as the rule evaluator, while GPT-6 Sol~\cite{openai2026gpt6sol} serves as the planning backbone.
We supply the evaluator with the stitch craft rules and each method's predicted plan.
The evaluator checks the assignment of each named object against the rules and the plan's style-level and stitch-budget metadata.
Let $c_o\in\{0,1\}$ indicate whether object $o$'s assignment is judged compliant under this rule system.
The score is
\begin{equation}
    \mathrm{RuleComp}
    =\frac{1}{|\mathcal{O}|}\sum_{o\in\mathcal{O}}c_o.
    \label{eq:supp_rule_comp}
\end{equation}
Rule Compliance evaluates the planning decisions against the specified knowledge base, while Stitch Accuracy evaluates their realization in generated images.


\subsection{User Study Details}
\label{sec:supp_user_study}


\paragraph{Expert Participants.}
We conduct the study with 10 Su embroidery experts, including next-generation inheritors and visual researchers, all with professional expertise in Su embroidery.
Drawing on diverse practical backgrounds, the inheritors assess the AI-generated results from the perspectives of craft execution and visual presentation.
The researchers contribute expertise in motif symbolism, intangible cultural heritage digitization, and visual culture, assessing image accuracy and adherence to craft conventions.
Considering both perspectives helps assess not only whether an output is visually convincing, but also whether its details remain consistent with established Su embroidery practices.
The combination of practical and theoretical expertise provides a balanced basis for the evaluation.

\begin{figure}[t]
    \centering
    \includegraphics[trim=172 200 193 115, clip, width=\columnwidth]{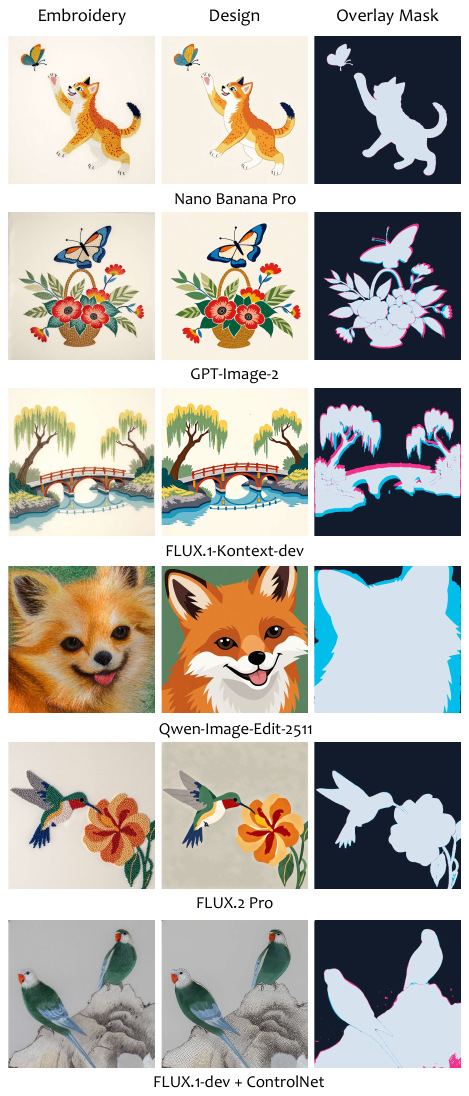}
    \caption{\textbf{Comparison of embroidery-to-design conversion methods.}
    Each row shows the source embroidery, converted flat design, and foreground-mask overlay.
    Colored contour differences visually indicate spatial drift.
    }
    \label{fig:reverse}
\end{figure}

\paragraph{Questionnaire Protocol.}
Each expert evaluates 30 randomly selected comparison groups. Each group presents one flat design, the requested style level, and the results of \methodname and one comparative method. Each candidate result includes its stitch-assignment plan and generated embroidery image. The presentation order of the results is randomized, and no model names are shown. Experts rate each candidate separately on the five criteria below using a five-point Likert scale, where 1 indicates \textit{strongly disagree} and 5 indicates \textit{strongly agree}. Higher scores indicate more favorable assessments. Scores are averaged separately for each method and each criterion. 
The questionnaire focuses on only the six stitch types studied: chain, plain, layered, knot, long-short, and crisscross.

\paragraph{Evaluation Criteria.}
\textbf{Aesthetics} evaluates the overall harmony, naturalness, and artistic appeal of the embroidery image.
\textbf{Realism} assesses how closely the generated textures resemble real silk-thread embroidery in luster, thread overlap, and layered appearance.
\textbf{Stitch Accuracy} measures whether the rendered stitch in each region matches its assigned type in the accompanying plan.
\textbf{Stitch Appropriateness} assesses whether the stitch choices follow traditional Su embroidery conventions and practical craft logic.
\textbf{Fabrication Feasibility} assesses whether the generated result is suitable as a reference for physical embroidery.

\section{Additional Experiments and Results}
\label{sec:supp_additional_experiments}

\subsection{Comparison of Embroidery-to-Design Conversion Methods}
\label{sec:supp_reverse_comparison}


\begin{figure}[t]
    \centering
    \includegraphics[trim=120 50 90 70, clip, width=\columnwidth]{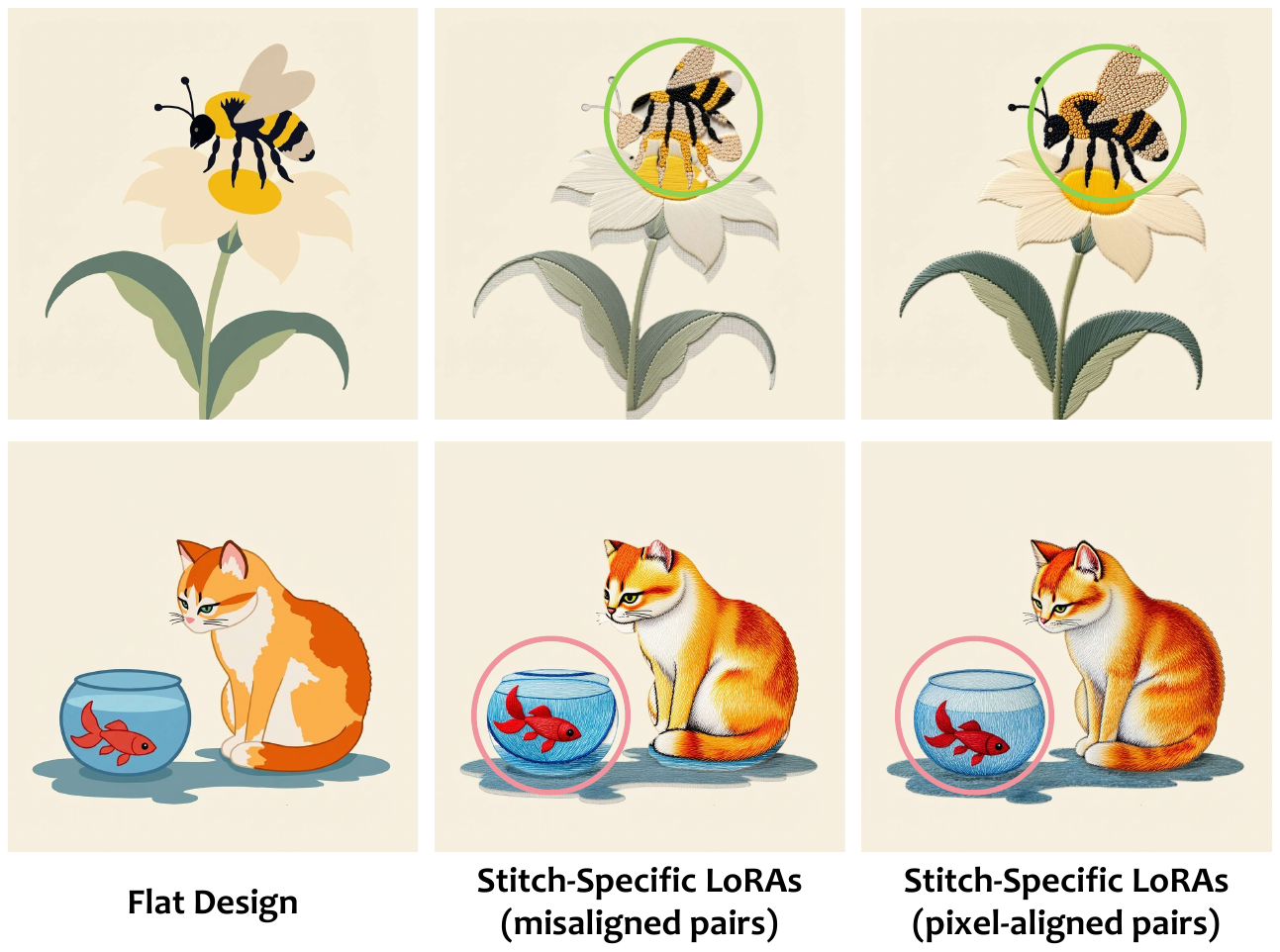}
    \caption{\textbf{Effect of training-pair alignment on stitch-specific rendering.}
    Left: input flat designs. Middle/right: results using stitch-specific LoRAs trained with misaligned/pixel-aligned design--stitch pairs.
    Colored circles highlight the bee and fishbowl regions, where misaligned supervision produces spatial drift. Pixel-aligned pairs better preserve the input positions and shapes.}
    \label{fig:misaligned}
\end{figure}

\begin{figure*}[t]
    \centering
    \includegraphics[trim=7 160 7 130, clip, width=\textwidth]{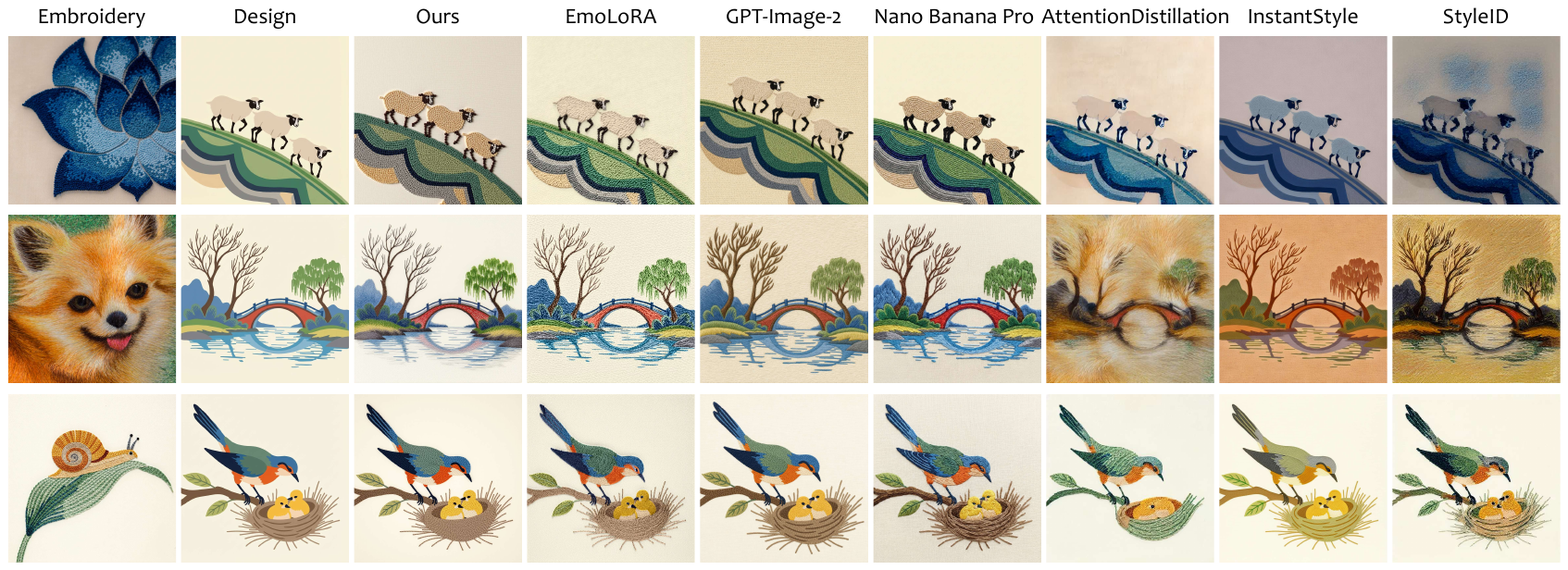}
    \caption{\textbf{Qualitative comparisons of global single-stitch embroidery synthesis.}
    Each row shows a target stitch reference, an input flat design, and results from our renderer, EmoLoRA, GPT-Image-2, Nano Banana Pro, Attention Distillation, InstantStyle, and StyleID.
    }
    \label{fig:qualitative_single}
\end{figure*}

\paragraph{Comparison Setting.}
The reverse-conversion model must remove thread texture while preserving the geometry and color organization needed for paired supervision.
We qualitatively compare Nano Banana Pro~\cite{google2025nanobananapro}, GPT-Image-2~\cite{openai2026gptimage2}, FLUX.1-Kontext-dev~\cite{batifol2025fluxkontext}, Qwen-Image-Edit-2511~\cite{qwen2025qwenimageedit2511}, FLUX.2 Pro~\cite{blackforestlabs2025flux2pro}, and FLUX.1-dev with ControlNet~\cite{blackforestlabs2024flux1,zhang2023controlnet}.
All methods receive the same embroidery inputs and instruction:
\begin{quote}
    \small\normalfont\itshape\raggedright
    ``Turn the embroidery artwork into a smooth digital painting with flat colors, retaining the structure and colors, removing all texture.''
\end{quote}
The comparison examines spatial correspondence and color preservation in the converted designs.
\textit{Spatial drift} denotes visible changes in a source motif's position or extent that break its alignment with the original embroidery.

\begin{table*}[t]
    \centering
    \caption{\textbf{Quantitative comparisons on global single-stitch embroidery synthesis.}
    Best and second-best results are highlighted in bold and underlined, respectively.
    LPIPS, Histogram Loss, and KID are scaled by $100$; HFRD is scaled by $1000$.
    }
    \label{tab:single_comparison}
    \small
    \setlength{\tabcolsep}{4pt}
    \renewcommand{\arraystretch}{1.15}
    \begin{tabular*}{\textwidth}{@{\extracolsep{\fill}}lccccccc@{}}
        \toprule
        Metric
        & Ours
        & EmoLoRA
        & GPT-Image-2
        & Nano Banana Pro
        & Attention Distillation
        & InstantStyle
        & StyleID \\
        \midrule
        LPIPS $\downarrow$
        & \textbf{19.839}
        & \underline{20.849}
        & 26.759
        & 29.663
        & 31.143
        & 23.099
        & 39.790 \\
        Hist. Loss $\downarrow$
        & 61.729
        & \textbf{54.339}
        & \underline{61.728}
        & 64.519
        & 77.636
        & 85.812
        & 82.535 \\
        KID $\downarrow$
        & \textbf{3.253}
        & \underline{4.931}
        & 5.070
        & 5.036
        & 5.154
        & 8.182
        & 7.278 \\
        HFRD $\downarrow$
        & \textbf{1.773}
        & \underline{6.606}
        & 7.553
        & 8.212
        & 7.253
        & 12.396
        & 9.828 \\
        \bottomrule
    \end{tabular*}
\end{table*}

\paragraph{Qualitative Observations and Model Choice.}
As shown in \Cref{fig:reverse}, GPT-Image-2, FLUX.1-Kontext-dev, and Qwen-Image-Edit-2511 exhibit spatial drift in the converted designs.
Such misalignment compromises the design--stitch correspondence used for supervision and can propagate spatial drift to renderers subsequently fine-tuned on these pairs; we therefore exclude these three models from pair construction.
Qwen-Image-Edit-2511 also shows limited color preservation.
FLUX.2 Pro and FLUX.1-dev with ControlNet avoid visible spatial drift in the shown examples, but FLUX.2 Pro alters background colors, while FLUX.1-dev with ControlNet produces less natural-looking conversions.
For this task, we select Nano Banana Pro because it preserves spatial correspondence without visible drift and retains the source colors well.
Expert filtering remains part of the construction pipeline after reverse conversion.

\paragraph{Effect of Pair Alignment on Renderer Training.}
\Cref{fig:misaligned} compares results obtained with stitch-specific LoRAs trained on misaligned and pixel-aligned design--stitch pairs.
Although a misaligned pair can depict the same subject, its design and embroidery disagree in pattern position, shape, or extent.
The highlighted bee and fishbowl show how such discrepancies can propagate to the learned transformation as spatial drift.
In contrast, pixel-aligned supervision better preserves the input geometry in these examples.
This correspondence is important for multi-stitch synthesis because composition uses masks defined on the input design: misaligned rendering results can no longer match their assigned regions.
We therefore treat structure alignment as a key criterion during pair curation and expert filtering.

\subsection{Comparison of Stitch-Specific Renderers}
\label{sec:supp_renderer_comparisons}


We additionally evaluate our stitch-specific renderers on global single-stitch embroidery synthesis.
This experiment isolates stitch-specific rendering; it does not involve craft-aware planning or area-ordered composition.

\paragraph{Compared Methods.}
We compare with general-purpose image editing models, reference-based style-transfer methods, and the embroidery-specific EmoLoRA.
\textbf{EmoLoRA}~\cite{ma2025emolora} targets one-shot embroidery customization through two-stage contrastive LoRA modulation to separate fine-grained embroidery style from content.
In our global single-stitch setting, it learns an embroidery appearance from a single stitch reference and applies the learned appearance to the entire input design.
\textbf{GPT-Image-2}~\cite{openai2026gptimage2} and \textbf{Nano Banana Pro}~\cite{google2025nanobananapro} are general-purpose image generation and editing models. In this task, they receive the flat design and target stitch reference to produce a globally stylized embroidery image.
\textbf{InstantStyle}~\cite{wang2024instantstyle} separates reference style and content in feature space and injects reference features into style-specific diffusion blocks.
\textbf{Attention Distillation}~\cite{zhou2025attentiondistillation} transfers reference visual characteristics using an attention-distillation objective based on diffusion self-attention features.
\textbf{StyleID}~\cite{chung2024styleinjection} is a training-free diffusion style-transfer method that injects reference style features while seeking to retain the content structure.
We also considered \textbf{MSEmbGAN}~\cite{hu2025msembgan}, another embroidery synthesis method, but could not include it in our comparisons because its implementation was not publicly available at the time of evaluation.


\paragraph{Qualitative Comparisons.}
\Cref{fig:qualitative_single} compares design preservation and target-stitch reproduction.
Our renderers retain the scene layout, object boundaries, and major color regions while reproducing distinct stitch textures.
EmoLoRA also preserves the input structure and colors well, demonstrating the value of learning a global embroidery appearance.
However, its one-shot adaptation does not reliably capture the characteristic textures of Su embroidery stitches.
Even knot stitch, whose raised, granular appearance is particularly distinctive, is not reproduced well; the other stitch types likewise show limited texture fidelity.
Other competing methods may change the palette, alter local structures, or transfer content from the stitch reference.
For example, the landscape case shows pronounced color changes in some style-transfer results and reference-content leakage in the Attention Distillation result.
These observations also suggest that global embroidery stylization alone is insufficient for this task.
Reliable synthesis must jointly preserve the input design and reproduce the local thread morphology characteristic of each target stitch.

\begin{figure}
    \centering
    \includegraphics[trim=150 65 145 60, clip, width=\columnwidth]{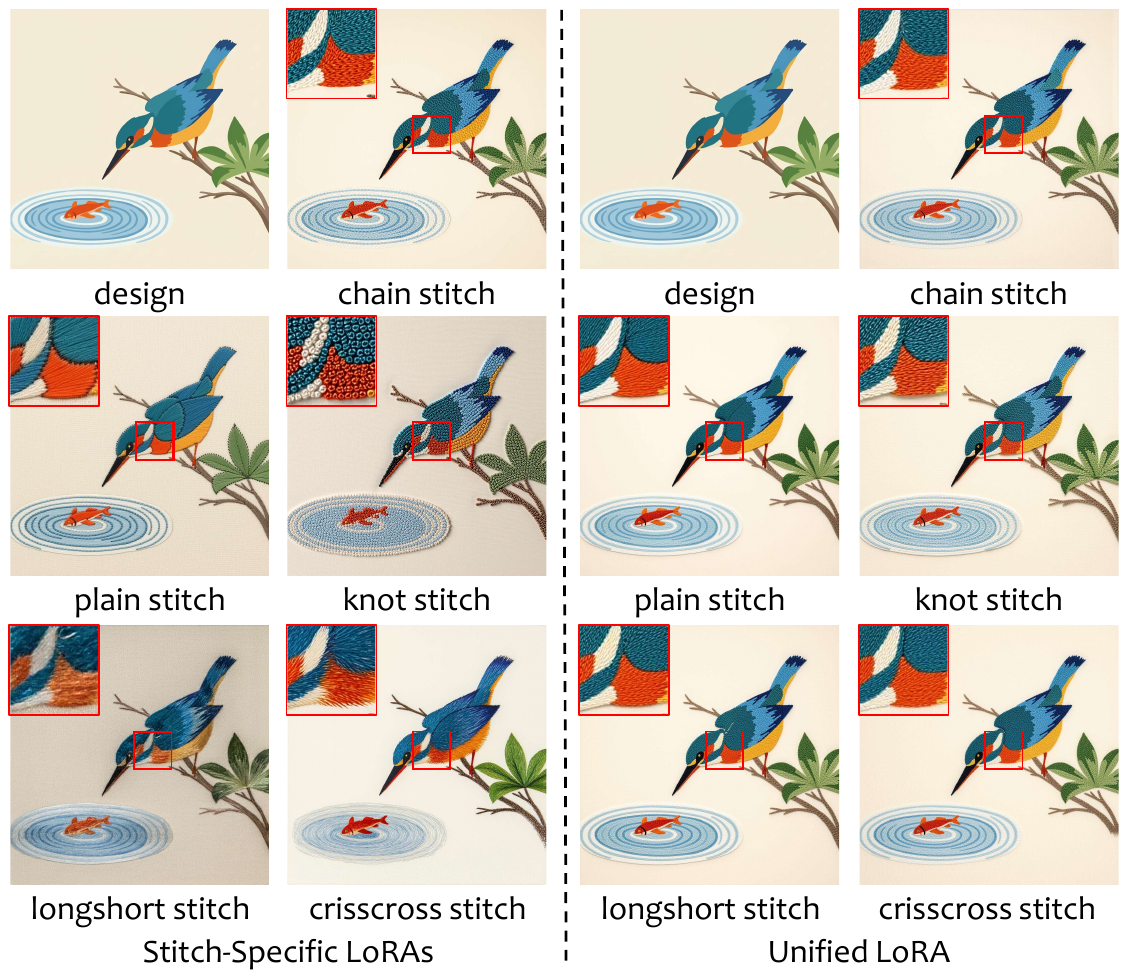}
    \caption{\textbf{Stitch-specific LoRAs versus a unified LoRA.}
    Separate LoRAs produce distinct stitch appearances, whereas the unified adapter produces similar thread patterns across prompts.
    Red boxes and enlarged insets highlight the local differences.}
    \label{fig:unifiedLoRA}
\end{figure}

\paragraph{Quantitative Comparisons.}
\Cref{tab:single_comparison} reports four complementary metrics for the global single-stitch task.
Our renderers achieve the lowest LPIPS ($19.839$), KID ($3.253$), and HFRD ($1.773$).
EmoLoRA ranks second on these three metrics, with LPIPS of $20.849$, KID of $4.931$, and HFRD of $6.606$, and achieves the lowest Histogram Loss ($54.339$).
Its strong perceptual and color-distribution scores are consistent with the good design preservation observed qualitatively.
Our Histogram Loss of $61.729$ ranks third, closely following GPT-Image-2 ($61.728$).
The lower KID and HFRD of our renderers indicate closer agreement with the target embroidery texture distributions, consistent with the clearer stitch morphology in \Cref{fig:qualitative_single}.
Together, these results support learning stitch-specific renderers from pixel-aligned design--stitch pairs to preserve design content while reproducing characteristic stitch textures.


\begin{figure*}[t]
    \centering
    \includegraphics[trim=115 70 120 70, clip, width=\textwidth]{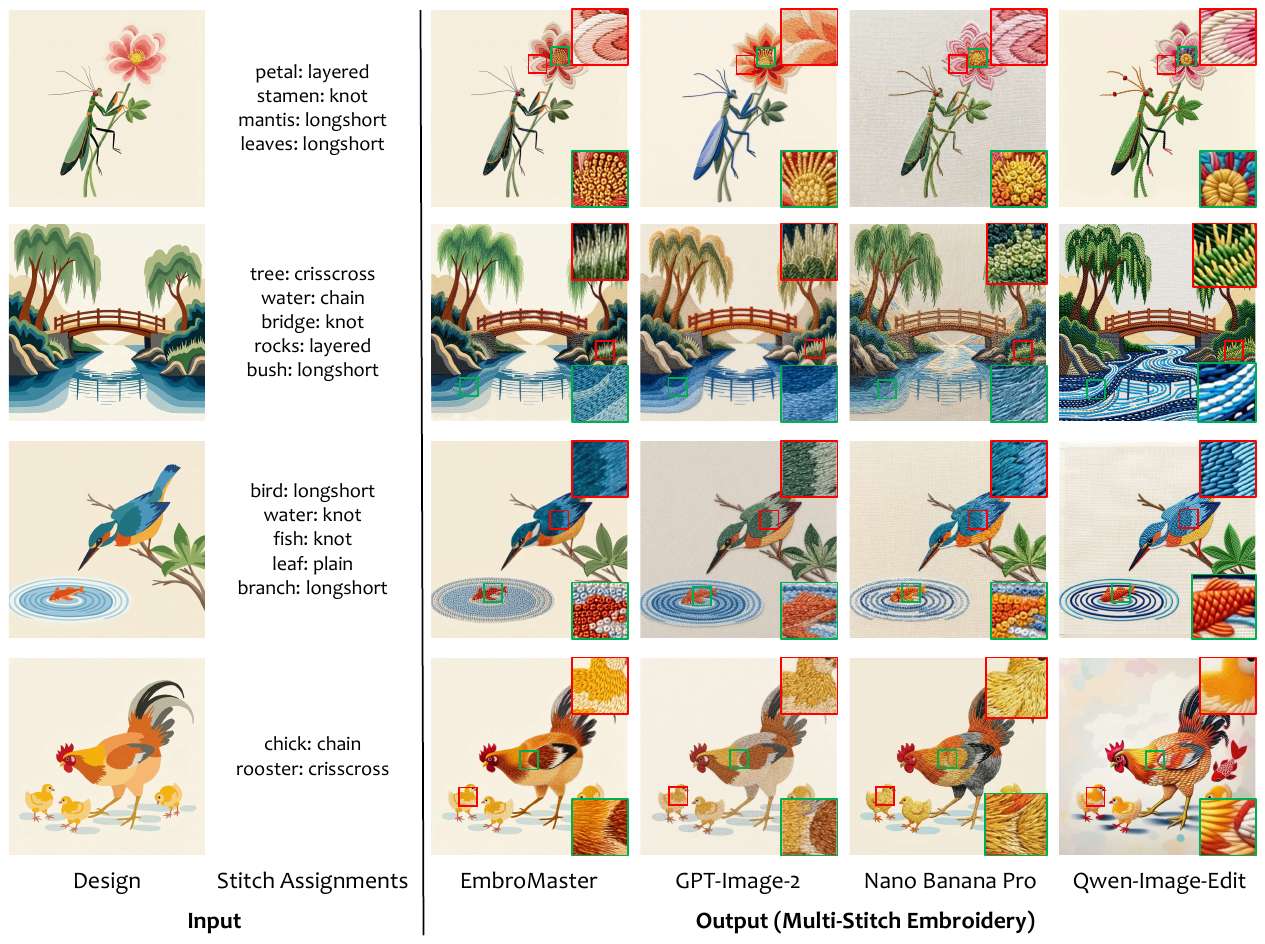}
    \caption{\textbf{Qualitative comparisons of multi-stitch synthesis with fixed stitch assignments.}
    Each row shows an input flat design, the common stitch assignments, and results from \methodname, GPT-Image-2, Nano Banana Pro, and Qwen-Image-Edit-2511.
    }
    \label{fig:qualitative_suppl}
\end{figure*}

\paragraph{Unified LoRA for Stitch Rendering.}
\label{sec:supp_unified_lora}
We investigate whether a single prompt-conditioned LoRA can replace the separate stitch-specific adapters.
Following the image-to-image training procedure in \Cref{sec:supp_renderer_training}, we fine-tune FLUX.1-Kontext-dev using the pooled design--stitch pairs from all six stitch types.
The joint dataset contains 20 pairs per stitch, for 120 pairs in total, and each pair retains its stitch-specific editing instruction.
We train one unified LoRA with rank 64, alpha 64, and a learning rate of $1\times10^{-4}$ for 6,000 steps.
The target stitch is specified through the same instruction template:
\begin{quote}
    \small\normalfont\itshape\raggedright
    ``convert to xxx\_stitch embroidery style''
\end{quote}
For example, \texttt{chain\_stitch} requests a chain-stitch rendition, while \texttt{layered\_stitch} requests a layered-stitch rendition of the input design.
The key distinction from our stitch-specific training is that all stitch types share a single adapter and are distinguished through their prompts.

\Cref{fig:unifiedLoRA} compares the unified LoRA with our stitch-specific renderers on the same input design and requested stitch types.
The unified LoRA exhibits confusion between stitch types and does not consistently reproduce the requested stitch morphology.
In the illustrated example, its outputs have similar thread patterns across different stitch prompts.
Even for knot stitch, whose granular appearance is visually distinctive, some outputs correctly exhibit knot-like structures while others resemble different stitch types despite being conditioned on the knot-stitch instruction.
The other target stitches show less reliable results and more obvious stitch confusion in the observed examples.
These qualitative findings indicate that prompt conditioning alone does not reliably separate the stitch appearances under this joint-training configuration.
The observed errors are consistent with interference between stitch types learned by a shared adapter, supporting our use of stitch-specific renderers for more reliable stitch control in the current framework.

\subsection{Multi-Stitch Synthesis with Fixed Stitch Assignments}
\label{sec:supp_fixed_assignments}


\paragraph{Experimental Setup.}
To evaluate multi-stitch generation independently of differences in planning, we use the same 200 flat designs as in the main paper and fix the stitch assignments for each design across all methods.
We compare \methodname with Qwen-Image-Edit-2511, GPT-Image-2, and Nano Banana Pro.
The baseline generators receive the common assignments through the synthesis prompt in \Cref{fig:prompt3}, while \methodname uses the corresponding stitch-specific renderers and area-ordered composition.
This experiment evaluates each method's ability to render a prescribed combination of stitches while preserving the design.
By holding the object-to-stitch mapping constant, the comparison eliminates planning quality as a confounding factor and isolates differences in the rendering strategies.
It further examines whether each method can reproduce the prescribed stitch textures within their assigned regions without disrupting the overall composition.

\begin{table*}[t]
    \centering
    \caption{\textbf{Quantitative comparisons of multi-stitch synthesis with fixed stitch assignments.}
    Best and second-best results are highlighted in bold and underlined, respectively.
    LPIPS, RSP, Hist. Loss, and L-KID are scaled by $100$; L-HFRD by $1000$; L-$\Delta E_{00}$ is unscaled.}
    \label{tab:qualitative_suppl}

    \small
    \setlength{\tabcolsep}{3pt}
    \renewcommand{\arraystretch}{1.15}

    \begin{tabular*}{\textwidth}{
        @{\extracolsep{\fill}}lccccccc@{}
    }
        \toprule
        Method
        & LPIPS $\downarrow$
        & RSP $\uparrow$
        & Hist. Loss $\downarrow$
        & L-$\Delta E_{00}$ $\downarrow$
        & L-HFRD $\downarrow$
        & L-KID $\downarrow$
        & Stitch Acc. $\uparrow$ \\
        \midrule

        Qwen-Image-Edit
        & 55.119
        & 35.296
        & 77.398
        & 26.619
        & 8.658
        & 10.808
        & 0.210 \\

        GPT-Image-2
        & 33.242
        & 37.842
        & 75.005
        & 27.601
        & 9.577
        & 10.538
        & \underline{0.352} \\

        Nano Banana Pro
        & \underline{25.184}
        & \underline{42.949}
        & \underline{56.279}
        & \underline{18.764}
        & \underline{8.560}
        & \underline{9.437}
        & 0.339 \\

        EmbroMaster
        & \textbf{10.508}
        & \textbf{61.105}
        & \textbf{31.212}
        & \textbf{13.927}
        & \textbf{3.985}
        & \textbf{8.138}
        & \textbf{0.602} \\

        \bottomrule
    \end{tabular*}
\end{table*}

\paragraph{Evaluation Metrics.}
We report LPIPS, Region-wise Structure Preservation (RSP), Histogram Loss (Hist. Loss), Local Color $\Delta E_{00}$ (L-$\Delta E_{00}$), Local High-Frequency Ratio Difference (L-HFRD), Local KID (L-KID) and Stitch Classification Accuracy (Stitch Acc.).
LPIPS and RSP assess perceptual and structural design preservation; Hist. Loss and L-$\Delta E_{00}$ assess color fidelity; L-HFRD and L-KID compare stitch-conditioned texture statistics and feature distributions with real stitch references; Stitch Acc. evaluates whether the rendered stitches match the stitch assignments.
Stitch Assignment Rule Compliance is not included because the stitch plan is shared across methods.

\paragraph{Qualitative Comparisons.}
\Cref{fig:qualitative_suppl} reveals substantial differences in stitch texture fidelity and adherence to the fixed stitch assignments. Qwen-Image-Edit-2511 produces the weakest results, exhibiting largely homogeneous textures across nearly all examples, with limited recognizable stitch-specific morphology and little resemblance to authentic embroidery threads. GPT-Image-2 generates more discernible embroidery-like textures, but their quality remains relatively low and often fails to capture the characteristics of the prescribed stitches. Nano Banana Pro produces more convincing stitch textures than GPT-Image-2, yet all three baselines exhibit inconsistencies between the rendered textures and the supplied stitch assignments. For example, in the second row, the bush is assigned long-short stitch, but Nano Banana Pro renders a granular texture resembling knot stitch. In the third row, both the fish and water are assigned knot stitch, whereas GPT-Image-2 fails to reproduce the characteristic raised granular structures. In contrast, EmbroMaster consistently produces distinct stitch-specific textures that closely match the prescribed assignments. Meanwhile, it better preserves the input design's geometry and color organization. These observations demonstrate EmbroMaster's ability to faithfully realize different stitch types within their designated regions, rather than merely generating a generic embroidery appearance.

\paragraph{Quantitative Comparisons.}

\Cref{tab:qualitative_suppl} shows that EmbroMaster achieves the best performance across all seven metrics, demonstrating superior design preservation, stitch texture fidelity, and adherence to the fixed stitch assignments. Specifically, EmbroMaster achieves the lowest LPIPS ($10.508$) and highest RSP ($61.105$), substantially outperforming the second-best Nano Banana Pro ($25.184$ and $42.949$, respectively), indicating better perceptual and structural preservation of the input designs. It also obtains the lowest Histogram Loss ($31.212$) and L-$\Delta E_{00}$ ($13.927$), reflecting stronger color fidelity, together with the lowest L-HFRD ($3.985$) and L-KID ($8.138$), indicating closer agreement with real embroidery texture statistics and feature distributions. While Nano Banana Pro ranks second on these six metrics, it achieves only the third-highest Stitch Acc. ($0.339$), slightly below GPT-Image-2 ($0.352$). In contrast, EmbroMaster achieves the highest Stitch Acc. ($0.602$), substantially outperforming all baselines. This distinction indicates that visually plausible embroidery textures do not necessarily correspond to the prescribed stitch types. With identical stitch assignments supplied to all methods, these results demonstrate that EmbroMaster more effectively combines design fidelity, realistic stitch texture synthesis, and accurate realization of object-level stitch assignments, independently of differences in stitch planning.

\subsection{Additional Results of \methodname}
\label{sec:supp_results}

\begin{figure*}[t]
    \centering
    \includegraphics[width=\textwidth]{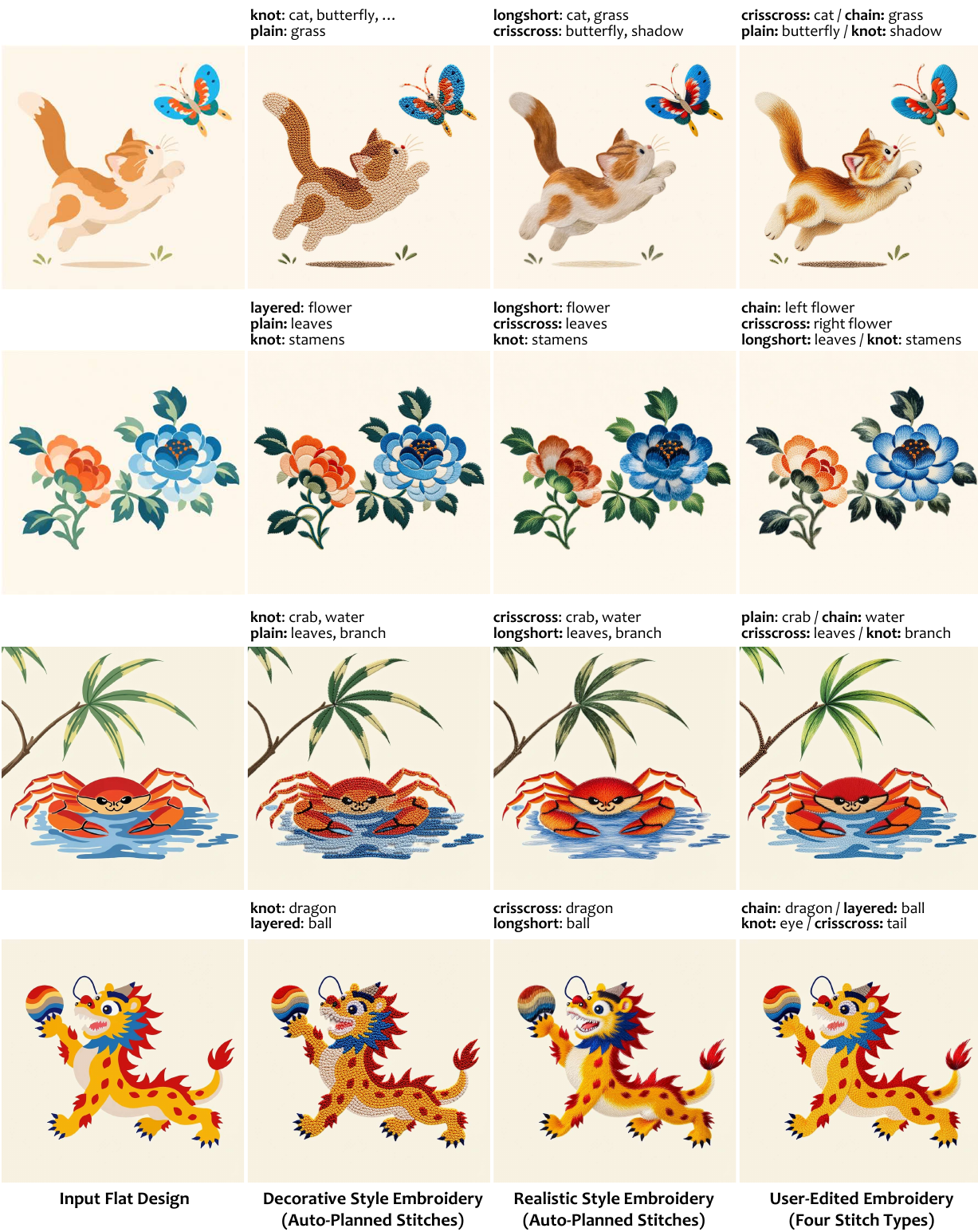}
    \caption{\textbf{Additional results generated by \methodname.}
    }
    \label{fig:our1}
\end{figure*}

\begin{figure*}[t]
    \centering
    \includegraphics[width=\textwidth]{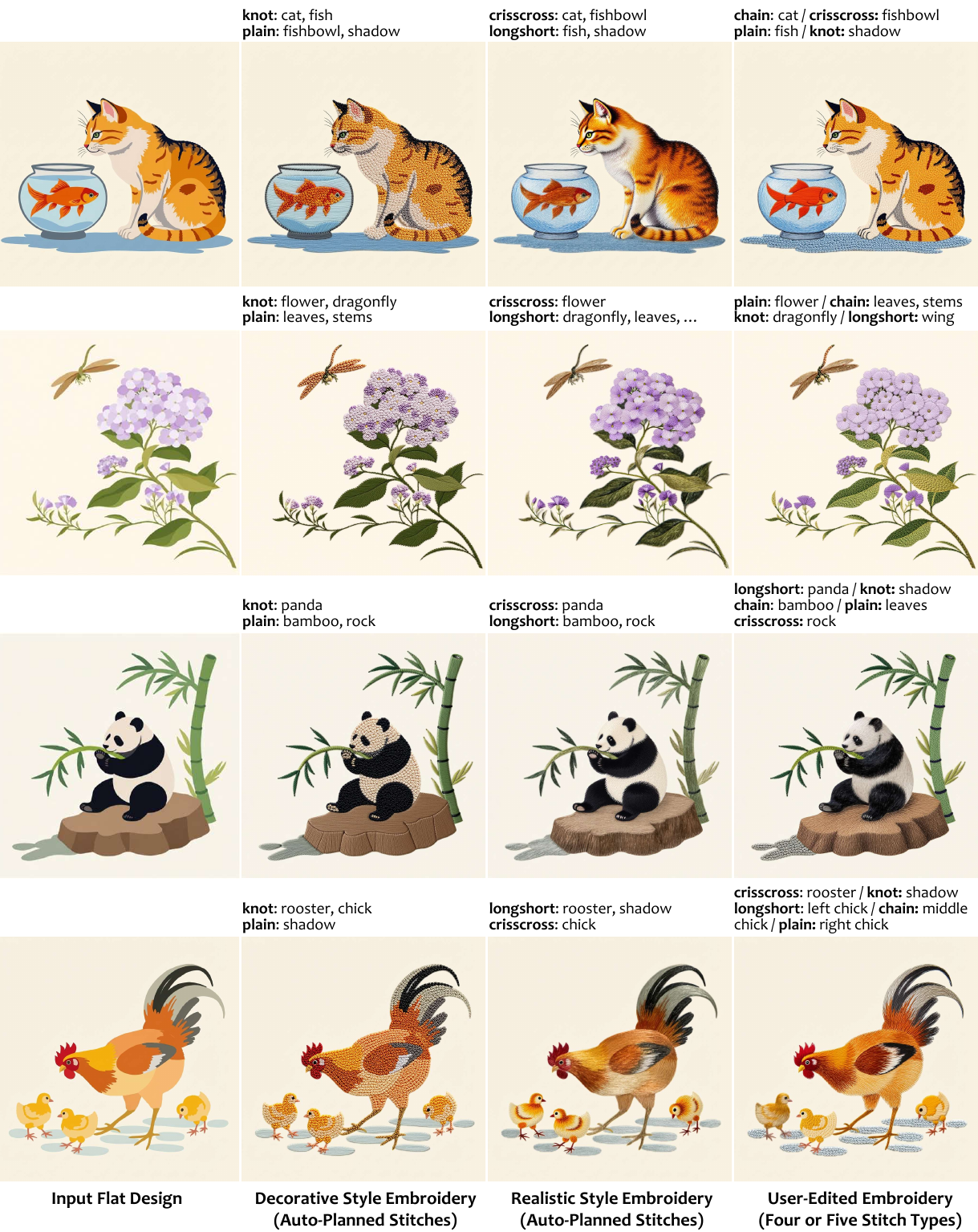}
    \caption{\textbf{Additional results generated by \methodname.}
    }
    \label{fig:ours2}
\end{figure*}

\begin{figure*}[t]
    \centering
    \includegraphics[width=\textwidth]{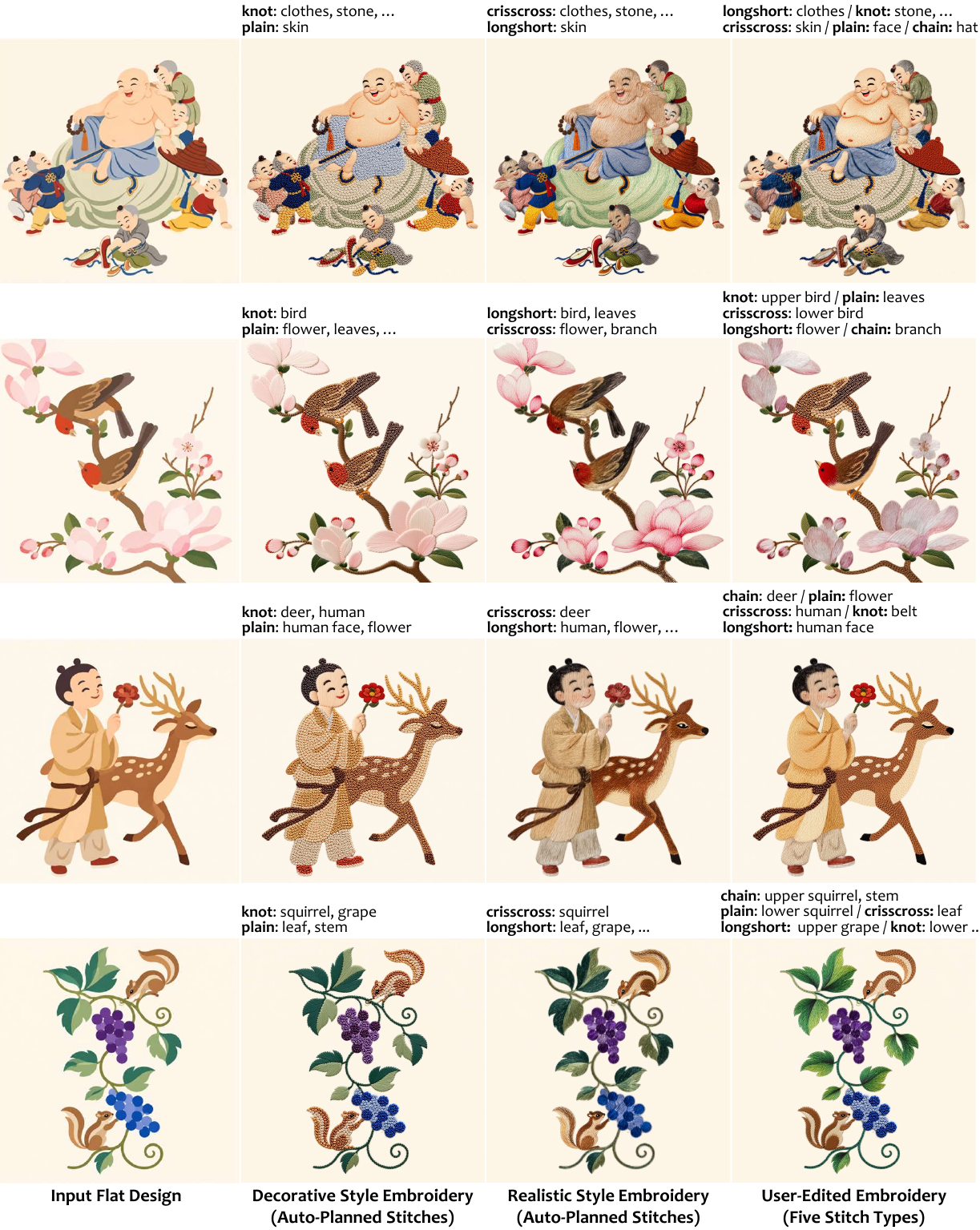}
    \caption{\textbf{Additional results generated by \methodname.}
    }
    \label{fig:ours3}
\end{figure*}

\begin{figure*}[t]
    \centering
    \includegraphics[width=\textwidth]{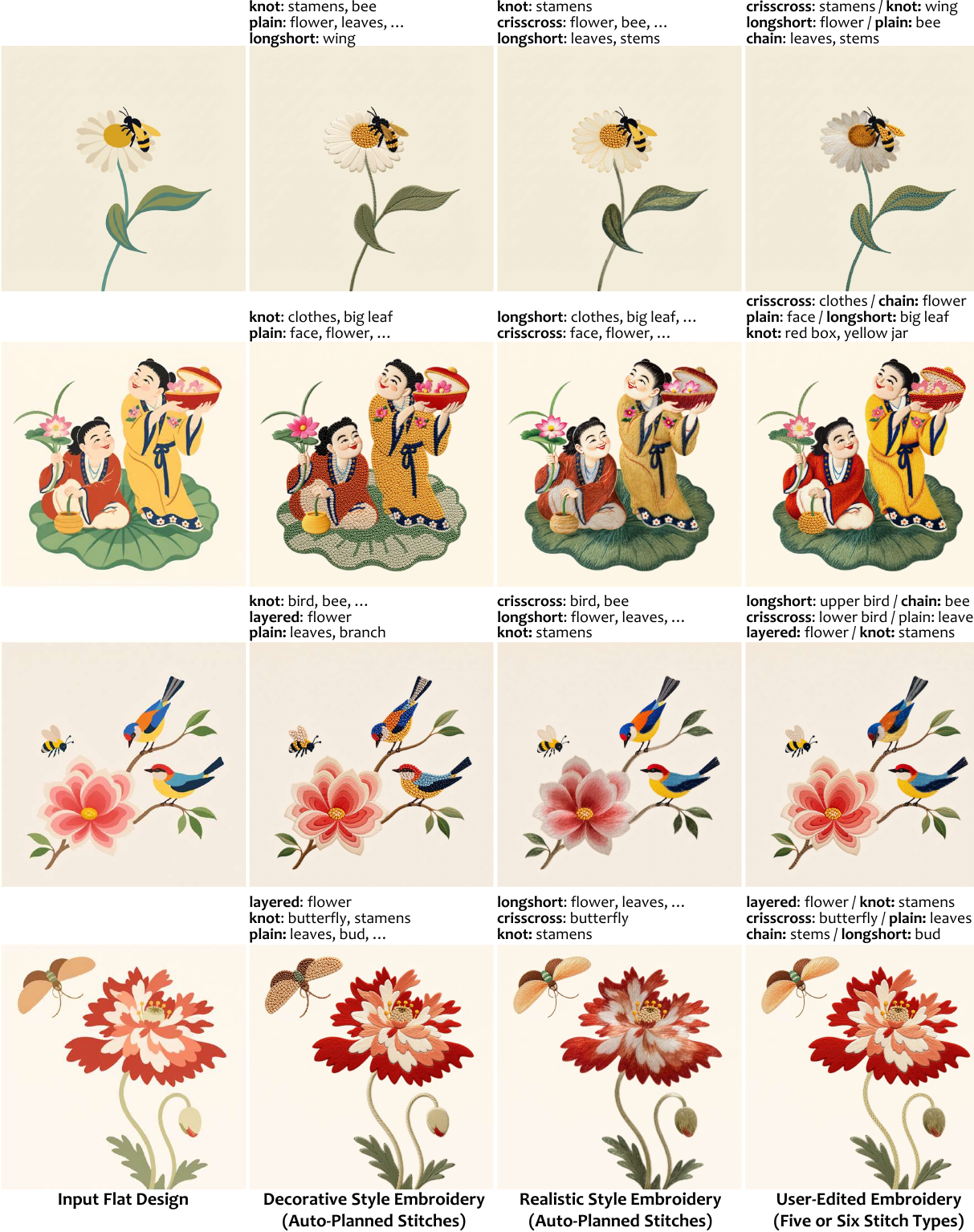}
    \caption{\textbf{Additional results generated by \methodname.}
    }
    \label{fig:ours4}
\end{figure*}

\Cref{fig:our1,fig:ours2,fig:ours3,fig:ours4} show additional results of EmbroMaster on a range of input designs. The examples cover floral patterns, animals, plants, human figures, and traditional dragon motifs. They range from simple designs with a few objects to detailed compositions with multiple subjects and small decorative elements. Each row presents the same design with different stitch assignments. The first column shows the input flat design. The second and third columns show decorative and realistic embroideries, respectively, with stitches automatically assigned by our craft-aware planning agent. The fourth column shows user-edited results, where users choose stitch types for specific objects or regions. These edited examples combine four to six stitch types in a single image, extending beyond the two- and three-stitch settings evaluated in the main paper. Users can also choose combinations beyond the automatic planning rules, such as combining chain stitch with other stitches. This allows users to explore different stitch combinations and customize the appearance of individual parts of the design.

\end{document}